\documentclass[final]{cvpr}

\usepackage{times}
\usepackage{epsfig}
\usepackage{graphicx}
\usepackage{amsmath}
\usepackage{amssymb}

\usepackage[dvipsnames]{xcolor}
\usepackage{bbm}
\usepackage{subfigure}
\usepackage{amsfonts}
\usepackage{enumerate}
\usepackage{enumitem}
\usepackage{tabu}
\usepackage{multirow}
\usepackage{makecell}
\usepackage{pifont}
\usepackage{color}
\newcommand{\cmark}{\ding{51}}%
\newcommand{\xmark}{\ding{55}}%
\usepackage{appendix}
\usepackage{float}
\usepackage{bm}
\newcommand{\tabincell}[2]{\begin{tabular}{@{}#1@{}}#2\end{tabular}}
\newcommand{\red}[1]{\textcolor{red}{#1}}
\newcommand{\green}[1]{\textcolor{green}{#1}}
\newcommand{\blue}[1]{\textcolor{blue}{#1}}

\colorlet{darky}{black!30!yellow!70!}
\colorlet{darkm}{black!20!magenta!80!}
\usepackage[pagebackref=true,breaklinks=true,colorlinks,bookmarks=false]{hyperref}

\def\cvprPaperID{5623} 

\begin{document}

\title{Higher Performance Visual Tracking with Dual-Modal Localization}

\setcounter{secnumdepth}{2}
\author{Jinghao Zhou\textsuperscript{\rm 1,2}, Bo Li\textsuperscript{\rm 2}, Lei Qiao\textsuperscript{\rm 2}, Peng Wang\textsuperscript{\rm 1}\\ 
Weihao Gan\textsuperscript{\rm 2}, Wei Wu\textsuperscript{\rm 2}, Junjie Yan\textsuperscript{\rm 2}, Wanli Ouyang\textsuperscript{\rm 3}\\
\textsuperscript{\rm 1}Northwestern Polytechnical University\\
\textsuperscript{\rm 2}SenseTime Group Limited\\
\textsuperscript{\rm 3}University of Sydney\\
\tt\small jensen.zhoujh@gmail.com, \{libo,qiaolei,ganweihao\}@sensetime.com,\\
\tt\small peng.wang@nwpu.edu.cn, wanli.ouyang@sydney.edu.au
}

\maketitle

\begin{abstract}
Visual Object Tracking (VOT) has synchronous needs for both robustness and accuracy.
While most existing works fail to operate simultaneously on both, we investigate in this work the problem of conflicting performance between accuracy and robustness. 
We first conduct a systematic comparison among existing methods and analyze their restrictions in terms of accuracy and robustness. Specifically, $4$ formulations-offline classification (OFC), offline regression (OFR), online classification (ONC), and online regression (ONR)-are considered, categorized by the existence of online update and the types of supervision signal.
To account for the problem, we resort to the idea of ensemble and propose a dual-modal framework for target localization, consisting of robust localization suppressing distractors via ONR and the accurate localization attending to the target center precisely via OFC.
To yield a final representation (i.e, bounding box), we propose a simple but effective score voting strategy to involve adjacent predictions such that the final representation does not commit to a single location. 
Operating beyond the real-time demand, our proposed method is further validated on $8$ datasets-VOT2018, VOT2019, OTB2015, NFS, UAV123, LaSOT, TrackingNet, and GOT-10k, achieving state-of-the-art performance.
\end{abstract}


\section{Introduction}
\label{sec:introductio}

Visual object tracking has synchronous needs for both robustness and accuracy. While the robustness stresses a tracker's discriminability against distractors, the accuracy focuses on precise localization of the optimal point to further derive a high-fidelity representation. An ideal design of tracking algorithm requires successful handling for both needs. 
In the literature, online trackers \cite{ECO,ATOM,KCF,DiMP} conduct online learning during tracking such that the weight for distinguishing target is adaptive to target deformation and object distraction. Though excelling in robustness by explicitly suppressing distractors, these approaches agonize over accurate tracking since regional center drift can be accumulated during online learning.
On the contrary, offline trackers \cite{SiamFC,DaSiamRPN,SiamRPN,SiamRPN++} exclude online update and locate the target via matching the template to its most similar area. While these approaches lag in their robustness against distractors due to an intrinsic insufficiency of constructing clear class boundaries, they are good at voiding disruption of accumulated error thus are accurate in target center localization.

\begin{figure}[!t]
\centering
\includegraphics[width=8.2cm]{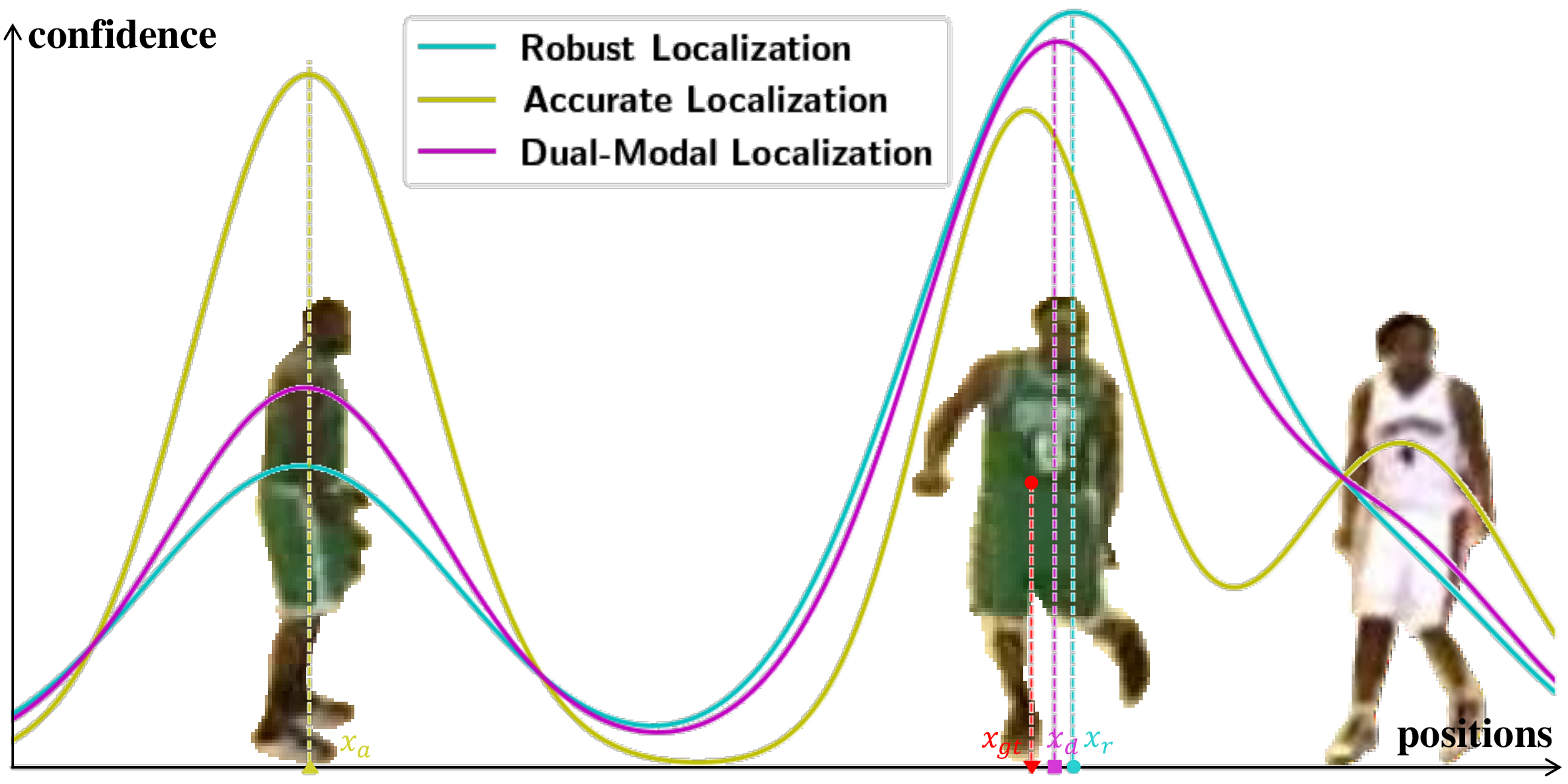}
\caption{An 1D schematic illustration of our motivation. \red{$x_{gt}$} is the groundtruth. The prediction of accurate localization \textcolor{darky}{$x_{a}$} is drifted severely but appears more shape-customized to derive boxes. The prediction of robust localization \textcolor{cyan}{$x_{r}$} handles well distinguishing distractors but fails to account for local details. The prediction of dual-modal localization \textcolor{darkm}{$x_{d}$} inherits both accuracy and robustness. 
}
\label{fig:firstfigure}
\vspace{-0.5cm}
\end{figure}

Considering that the requirements of robustness and accuracy can be contradictory, with the former insensitive to local details and robust in locating a rough target region while the latter sensitive to local details and accurate in determining optimal point for target representation, a divide-and-conquer strategy that solves them separately would be better than solving them integrally. Therefore, a dual-branch framework consisting of robust online tracking and accurate offline tracking arises naturally.

When robustness and accuracy are taken into further consideration, the choice of supervision signals matters.
For robust online tracking, a model trained with continuous Gaussian distribution can outperform that with a binary Bernoulli distribution, since target drift is more likely to happen when the binary Bernoulli distribution assigns all online predicted positives with equal significance. On the other hand, for accurate offline tracking, a Bernoulli distribution would be better with a more customized label assignment strategy \cite{ATSS,AutoAssign}, especially in the presence of an extra anchor-scale dimension for anchor-based trackers \cite{SiamRPN,SiamRPN++}. 
As shown in Figure \ref{fig:firstfigure}, a dual-modal localization framework can inherit both accuracy and robustness, contributing to higher performance visual tracking.

Another important topic in visual tracking is the regression of target representation (i.e, the bounding box) upon the localized target center. Compared with that in static images, such in the field of object detection \cite{YOLO,FastRCNN,FCOS}, box regression in tracking bear an intrinsic uncertainty of the target motion, deformation, and scale changes, etc. 
While recent studies \cite{IoUNet,ATOM} resorts to overlap maximization of randomly drawn boxes via gradient descent, enabling a quick adaption to the uncertainty of target shape, such approach eludes accurate box regression with potentially ill-reached solutions due to random initial states and requires extra computational cost. To account for the uncertainty intrinsic in target representation for tracking, a simple but effective score voting strategy is proposed and directly involves adjacent predictions around the target center, improving the tracker's capability for accurate tracking.

Our contributions can be summarised as three folds:  
\begin{itemize}
\vspace{-2mm}
\item We conduct a systematic comparison between $4$ major existing approaches-ONR, ONC, OFR, and OFC in terms of accuracy and robustness. We further analyze the intrinsic restrictions of their design.
\vspace{-2mm} 
\item We propose a dual-modal localization framework decoupling the task of target localization into robust localization via ONR and accurate localization via OFC.
\vspace{-6mm} 
\item We propose a score voting strategy to account for the uncertainty in target representation, which is achieved by involving box predictions around the target center. 
\vspace{-2mm}
\end{itemize}

Together with the dual-modal localization framework and score voting strategy, our methods achieve state-of-the-art performance under major benchmarks while operating at a real-time speed at $41$ Frame-Per-Second (FPS) on an NVIDIA Titan Xp GPU.
Results on major benchmarks: VOT2018 \cite{vot2018}, VOT2019 \cite{vot2019}, OTB2015 \cite{otb2015}, NFS \cite{nfs}, UAV123 \cite{uav123}, LaSOT \cite{lasot}, TrackingNet \cite{trackingnet}, and GOT-10k \cite{got10k} verify the effectiveness of our method.

\section{Related Work}
\label{sec:related}

The problem of visual tracking can be decomposed into two tasks: \textit{target localization} where we first locate a rough region of the target and \textit{target representation} where we further regress a high-fidelity representation of it. Next, we shall discuss detailedly how the two tasks have been previously solved by the existing trackers.

\noindent\textbf{Target Localization.} Several recent studies \cite{PrDiMP,EnergyModel} formulate the problem of \textit{target localization} as $y^\ast=\arg\max_y s_\theta(y,x)$, termed as \textit{confidence-based regression}, where $s_\theta:\mathcal{Y}\times\mathcal{X}\to\mathbb{R}$ predicts a scalar confidence score given an output-input pair $(y_i,x_i)$. To generate the localization heatmap convoluted by target-guided kernel formulated as $s_\theta(y,x)=(w * \phi(x))(y)$, the difference between the major trackers thus lies crucial in different paradigm in determining $w$.
While offline trackers (e.g, siamese trackers \cite{SiamFC,SiamRPN,SiamRPN++,SiamKPN}) set $w$ directly as the target feature $\phi(z)$ and keep it fixed during tracking, online trackers (e.g, correlation-filter trackers \cite{KCF,ECO} and classifier-based trackers \cite {MDNet,ATOM,DiMP,Det-MAML}, etc.) dynamically update $w$ with online generated pseudo labels.
Besides a categorization over tracker's update strategy, the distribution of pseudo label generated by the function $a: \mathcal{Y}\times\mathcal{Y}\to\mathbb{R}$ distinguishes the existing approaches into trackers based on classification and regression. For classification problem \cite{SiamFC,SiamRPN++,MDNet,Det-MAML}, the pseudo label $a(y,y_i) \in [0, 1]$ is binary (subject to Bernoulli distribution) and assigned often heuristically through Intersection-over-Union (IoU) between anchor and groundtruth boxes, with training loss $l(a(y,y_i), s_\theta(y,x_i))$ set to cross-entropy loss. The pseudo label for regression problem \cite{SiamKPN,DiMP,ATOM,KCF} is continuous between $(0, 1)$ and often subject to a Gaussian distribution concentrated on target center with training loss being generative loss, like L2 or hinge loss.
Following these two major categorizations of tracking algorithms, a combination of them results in $4$ major formulations to solve tracking problem: online regression (ONR), online classification (ONC), offline regression (OFR), and offline classification (OFC).


\noindent\textbf{Dual-Branch Structure for Target Localization.} Our proposed dual-modal framework is a dual-branch structure, which has been previously studied by several works. 
\cite{SA-Siam,Siam-BM} decouples two branches respectively accounting for appearance and semantic features to represent the tracked object. \cite{DROL,ANTI} consists of an online and an offline branch. While the offline branch is noise-free and allows no ambiguity, serving as guidance in selecting refined bounding boxes or masks, the online branch is rather discriminative against distractors. 
Our method resembles the last genres by decoupling into two localization models focusing respectively on robustness and accuracy. Different from them, which can be seen as a naive ensemble of two independent parts, our dual-modal framework achieves an end-to-end training and is validated to be optimal through a cross-over evaluation.

\noindent\textbf{Target Representation.} The target representation has been extensively studied in the field of visual tracking. Representations like bounding boxes \cite{ATOM,SiamRPN}, masks \cite{SiamMask}, keypoints \cite{RPT,SiamKPN}, or their combination \cite{AlphaRefine}) greatly boost the tracker's performance in precise tracking. In this work, we focus on the discussion of the bounding box as representation. The traditional paradigm follows the idea of \textit{direct regression} through $b_\theta: \mathcal{X}\to\mathcal{Y}$ by directly adjusting randomly-drawn boxes \cite{MDNet} or preset anchor boxes \cite{SiamRPN,SiamRPN++} around the coarsely localized target center. To tackle the commitment to a single state of \textit{direct regression}, a recently-proposed study \cite{ATOM} performs a gradient-descending optimization strategy via IoU maximization. In this work, we propose a simple but effective score voting strategy to involve the adjacent locations around the target center to entail uncertainty in target representation.

\section{Method}

The problem of visual tracking has two synchronous needs for both accuracy and robustness. However, most of the trackers are unable to handle well the two requirements simultaneously given the intrinsic restrictions of their design. In this work, we first give a comparison and analysis of existing uni-modal framework for \textit{target localization} in Section \ref{sec:analysis}. We then introduce the proposed dual-modal localization framework in Section \ref{sec:dualmodal}. Further, an effective score voting strategy for \textit{target representation} is proposed in Section \ref{sec:iouvoting}.

\begin{table}[!t]
\setlength{\tabcolsep}{0.2mm}{
\centering
\begin{tabu}{c|c|c|c|c|c} 
form & method & update & distribution & A$^*\uparrow$ & R$^*\downarrow$ \\
\Xhline{1.2pt} 
ONR & DiMP \cite{DiMP} & \cmark & Gaussian & 0.582 & \color{red}{\textbf{0.126}} \\
\hline
ONC-$1s$ & \multirow{2}{*}{Det-MAML \cite{Det-MAML}} & \multirow{2}{*}{\cmark} & Bernoulli-$1s$ & 0.577 & 0.136 \\
ONC-$5s$ & & & Bernoulli-$5s$ & 0.587 & 0.133 \\
\hline
OFR & SiamKPN \cite{SiamKPN} & \xmark & Gaussian & 0.597 & 0.284 \\
\hline
OFC-$1s$ & SiamCAR \cite{SiamCAR} & \multirow{2}{*}{\xmark} & Bernoulli-$1s$ & 0.599 & 0.262 \\
OFC-$5s$ & SiamRPN++ \cite{SiamRPN++} & & Bernoulli-$5s$ & \color{red}{\textbf{0.603}} & 0.254 \\
\end{tabu}
}
\setlength\abovecaptionskip{0cm}
\caption{Comparison of $4$ major formulations with their typical work, update strategy, label distribution, A (accuracy) and R (robustness). $^*$ denotes the average results over multiple sets of hyper-parameters to ensure stability. See Appendix \iftoggle{cvprfinal}{\ref{app:cumulative}}{\iftoggle{cvprrebuttal}{}{C}} for details.}
\label{tab:comparison}
\vspace{-0.3cm}
\end{table}

\begin{figure}[!t]
\centering
\includegraphics[width=7cm]{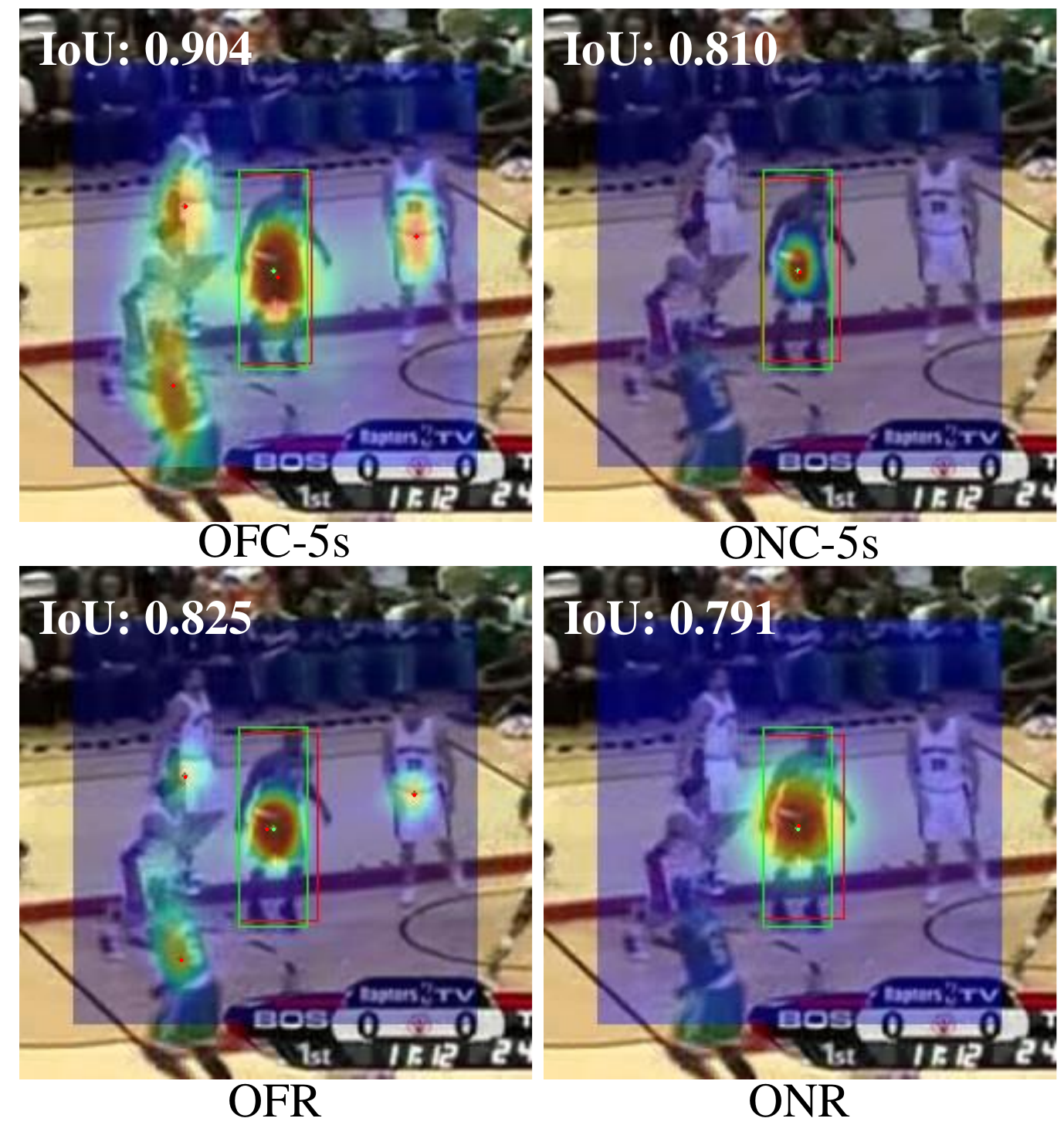}
\caption{Accuracy vs. Robustness. \green{Green} box and point denote groundtruth box and center. \red{Red} box and points denote predicted box and local peaks. IoU between the prediction and groundtruth boxes are shown. Best viewed with color and zooming in.}
\label{fig:analysis}
\vspace{-0.5cm}
\end{figure}

\begin{figure*}[!t]
\centering
\includegraphics[width=16cm]{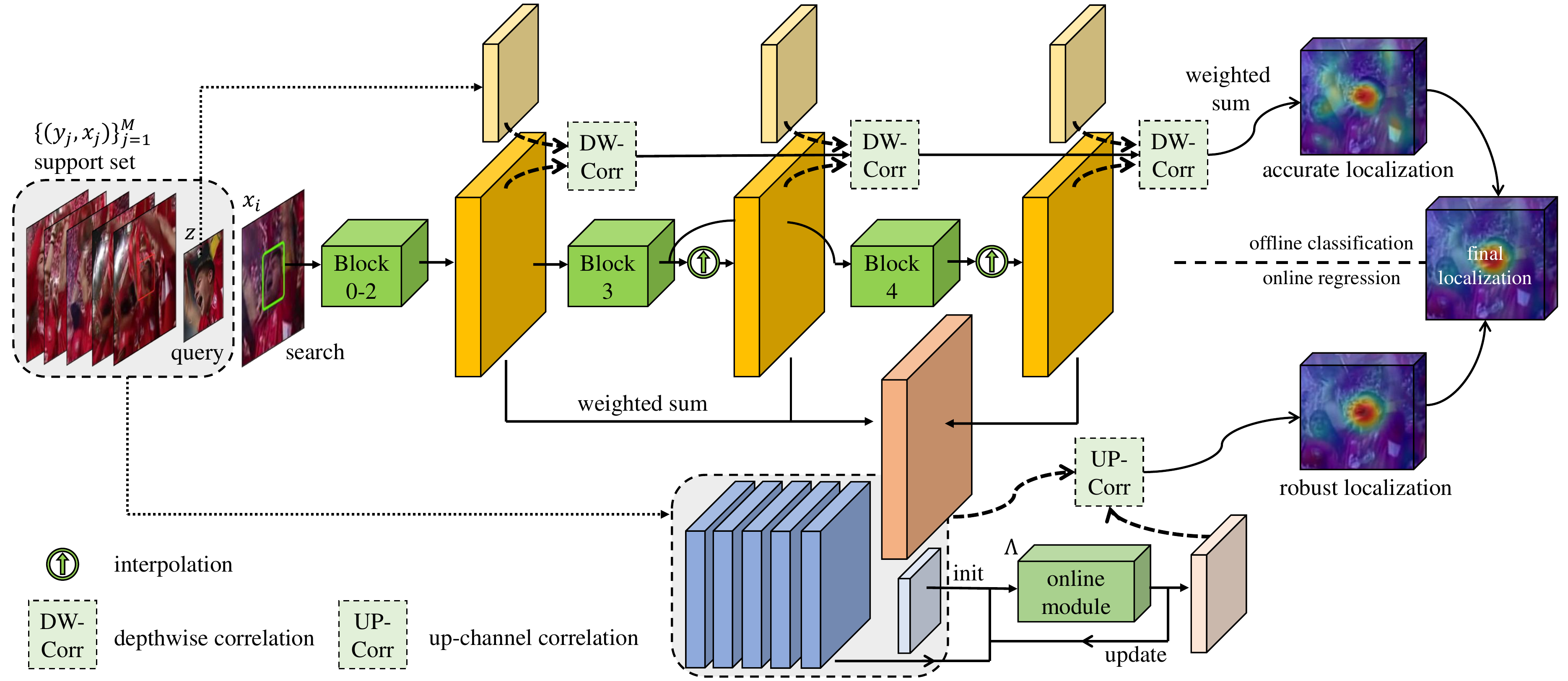}
\caption{Architecture of proposed dual-modal localization framework. While OFC is applied for accurate localization, ONR is applied for robust localization. The final localization is weighted-sum of the two models. The derivation of bounding box is left out for succinctness.}
\label{fig:architecture}
\vspace{-0.5cm}
\end{figure*}

\subsection{Accuracy vs. Robustness}
\label{sec:analysis}

In this section, we first conduct a comparison among the major existing methods in terms of accuracy and robustness. Specifically, the Accuracy (A) and Robustness (R) proposed in VOT challenge \cite{vot2018} are chosen, where A computes the average overlap of tracked frames and R computes the average ratio of lost frames in a sequence. 
Specifically, OFC, OFR, ONC, and ONR, categorized by the existence of update strategy and the types of supervision signals, are evaluated. 
While the offline tracker does not need further qualification, for online trackers, we focus on the discussion of recently proposed state-of-the-art \cite{DiMP} in a meta-learning fashion, which leverages gradient-based strategy within an inner-loop to allow an end-to-end training of the backbones. The classification is supervised by Bernoulli distribution while the regression is by Gaussian distribution.
For a fair comparison, we unify the box branch same as in \cite{SiamRPN++}. 
The results are shown in Table \ref{tab:comparison}. -$\{1,5\}s$ denotes that the additional anchor scale dimension for classification is applied. The implementation details can be found later in Section \ref{sec:apprachdetails}. Next, we give a theoretical analysis of the pros and cons of each method in the context of their A and R.

\noindent\textbf{Online vs. Offline.} 
From the Table \ref{tab:comparison}, we observe that one model often handles well one respect at a sacrifice of the other. \textit{Specifically, we find that the online approaches excel at robustness while the offline approaches excel at accuracy.} We first give a theoretical analysis of how the conflicting performance under two metrics is caused. The superiority of robustness of online approaches stems from
\begin{itemize}
\vspace{-2mm}
\item The online memory helps to store information about target historical deformation and potential distractors.
\vspace{-2mm} 
\item The weight update helps the fast and discriminative parameterization via optimization under L$2$ loss.
\vspace{-2mm}
\end{itemize}
On the contrary, offline approaches' excellence at accuracy can be ascribed to
\begin{itemize}
\vspace{-2mm}
\item The weight fixation avoids the ambiguity and accumulated error when assigning pseudo-labels.
\vspace{-2mm} 
\item The weight fixation avoids generating target-centered prediction and allows more shape-customized prediction suitable for box regression.
\vspace{-2mm}
\end{itemize}
For a visual demonstration, we provide the tracking results in Figure \ref{fig:analysis}. Online approaches focus only on target while offline approaches have multiple peaks, validating the superiority of online approaches on robustness. Also, offline approaches tend to give larger center shift between the prediction and groundtruth, resulting in higher IoU scores.
Next, we further investigate the influence of training labels on offline approaches and online approaches, focusing on the discussion of accuracy and robustness separately.

\noindent\textbf{OFC vs. OFR.} Comparing OFC-$1s$ and OFR, we find that OFC-$1s$ is slightly better in terms of accuracy. While the Gaussian distribution forces the adjacent region to have a lower score than the target center, the Bernoulli distribution allows multiple positives to be $1$. Therefore, OFC-$1s$ allows more uncertainty in deciding the optimal one to derive the final target representation.
Beyond that, OFC-$5s$ with an additional dimension of anchor scale further encodes representation information compared with single-scale regression or classification, thus achieves the optimal accuracy.

\noindent\textbf{ONC vs. ONR.} Comparing ONC-\{$1,5s$\} and ONR, ONR is slightly better than ONC in terms of robustness score. For robustness, we may expect the model to output a more target-centered prediction to avoid target drift. Therefore, the attribute of regression to suppress the surroundings around the target center can be beneficial in this regard. Further, compared with ONC, ONR allows more ambiguity in pseudo-label with wider peak region thus the surrounded pixels around target center remain certain probability to be considered as positive.

From the above analysis, we conclude that the advantages of specific formulation performing well under one metric also become the disadvantages of them performing well under the other, since uni-modal frameworks have their intrinsic restrictions of design. Intuitively, we consider a simple but effective ensemble of two diverse localization models that can handle well the separate needs of accuracy and robustness. 


\subsection{Dual-Modal Localization}
\label{sec:dualmodal}

In this section, a dual-modal localization framework decoupled into robust localization and accurate localization is proposed. Following the above experimental and theoretical analysis, we formulate robust localization into an ONR and accurate localization into an OFC. The two branches are trained independently with their own supervision signals and are directly fused to yield the final localization $\hat s_\theta$ by weighted sum during inference, formulated as
\begin{equation}
\label{form:fusion}
    \hat s_\theta(y,x_i) = \mu s^r_\theta(y,x_i) + (1 - \mu) s^a_\theta(y,x_i)
\end{equation}
where $\mu$ is a re-weighting hyper-parameter, stressing the significance of accurate and robust localization. Next, we shall detail how two localization models are implemented.

\noindent\textbf{Accurate Localization.} The accurate localization $s^a_\theta$ applies OFC thus resembles anchor-based SiamRPN++ \cite{SiamRPN++} by aggregating layer-wise correlation between first-frame template and the search region, formulate as
\begin{equation}
\label{form:siamese}
    s^a_\theta(y,x_i) = \sum_{l=3}^5 \alpha^l (\phi^l_\theta(z)*\phi^l_\theta(x_i))
\end{equation}
where $z$ is the template given by the first frame and $\phi$ is the network parameterized by $\theta$. The correlation results of each layer are weighted by $\alpha^l$, which is jointly trained with the network. Specifically, the classification considers anchor an extra scale dimension besides spatial dimension for more accurate localization, yielding a heatmap for accurate localization $s^a_\theta \in H\times W \times A$. Detailed modifications and differences with SiamRPN++ can be found in Section \ref{sec:apprachdetails}.

\noindent\textbf{Robust Localization.} In contrast, the robust localization $s^r_\theta$ applies ONR in a meta-learning fashion with inner-loop update. Out of the consideration of computation efficiency, we conduct the aggregation in feature level thus only one fused weight is updated instead of a set of weights for each layer. The robust localization can be derived as
\begin{equation}
\begin{aligned}
    s^r_\theta(y,x_i) &= \varphi * \sum_{l=3}^5 \beta^l \phi^l_\theta(x_i), \\
    \mathrm{where} \ \varphi &= \Lambda( \sum_{l=3}^5 \beta^l \phi^l_\theta(z), \{(\hat a(y,y_j), \sum_{l=3}^5 \beta^l \phi^l_\theta(x_j))\}^M_{j=1}; \rho)
\end{aligned}
\end{equation}
$\varphi$ is an online updated and sequence-specific weight and $\Lambda(\cdot;\rho)$ is an online module parameterized with $\rho$. For gradient-descent-based online module, $\rho$ can be learned parameters such as learning rate, step size, or online generated pseudo label. The online module $\Lambda$ takes fused feature of $z$ as initial weight and are updated using a online-maintained feature support set $\{(\hat a(y,y_j), \sum_{l=3}^5 \beta^l \phi^l_\theta(x_j))\}_{j=1}^M$ as online training samples. $\hat a(y,y_i)$ is online generated pseudo label and $\beta^l$ is learned weight for layer $l$ to aggregate the feature for fast online learning. By assigning Gaussian label for regression only on spatial scale, the heatmap for robust localization $s^r_\theta \in H\times W$ can be obtained. Its weighted sum with offline classification score are conducted by a naive broadcasting over the missing dimension. 

A complete architecture of our proposed dual-modal localization pipeline are shown in Figure \ref{fig:architecture}. After obtaining the final localization, the state of target center to further regress target's box representation is derived as $y^\ast=\arg\max_y \hat s_\theta(y,x_i)$.


\subsection{Score Voting for Box Representation}
\label{sec:iouvoting}

The siamese trackers with Regional Proposal Network (RPN) \cite{SiamRPN,SiamRPN++} obtain a final box representation by performing a direct regression upon the localized target center $y^\ast$ as $b^\ast = b_\theta(y^\ast,x_i)$ from the dense predictions $b_\theta(y,x_i)$ over spatial scale, which is obtained similar to Equation \ref{form:siamese} in an offline manner. However, the direct regression of box representation commits to only a single state $y^\ast$, ignoring potentially informative predictions on adjacent pixels. 
Inspired by \cite{KLLoss}, we acquire the final box representation $b^\ast$ via a score voting strategy, where adjacent predictions around $y^\ast$ from $b_\theta(y,x_i)$ are leveraged to account for target's multiple states and uncertainty in its shape, aspect ratio, and motion, etc. Our proposed score voting strategy can be formulated as
\begin{equation}
\begin{aligned}
     b^\ast &= \frac{\int w(y,x_i)o_\theta(y,x_i)b_\theta(y,x_i)dy}{\int w(y,x_i)o_\theta(y,x_i)dy} , \\
     \mathrm{where} \ y &\in \{y|\mathrm{IoU}(b_\theta(y,x_i),b^\ast) \textgreater \epsilon\} 
\end{aligned}
\end{equation}
where $o_\theta(y,x_i)$ is the IoU scores between the predicted box and groundtruth box via an additional IoU branch, the derivation of which can similarly resort to Equation \ref{form:siamese}. The IoU branch is trained end-to-end together with box branch and dual-modal localization. $w(y,x_i)$ is a penalized weight derived as
\begin{equation}
    w(y,x_i) = \hat s_\theta(y,x_i)\odot e^{-(1-\mathrm{IoU}(b_\theta(y,x_i), b^\ast))^2/\sigma}
\end{equation}
where $\hat s_\theta$ the final localization scores from Equation \ref{form:fusion}. The subsequent term represents a prior of Gaussian distribution of the real IoU between box representation at target center and adjacent box representations around. $\odot$ denotes point-wise multiplication and $\sigma$ is the variance term controlling the significance of the adjacent predictions and is set heuristically as a hyper-parameter. Compared with recently proposed \cite{ATOM}, which applies a gradient-descent-based strategy to account for multiple states and uncertainty of target's box representation, the obtaining of final box representation $b^\ast$ via our score voting strategy is entirely fully-feedforward and introduce minimal extra computation cost to the current network. We provide a visualized demonstration to show the effectiveness of proposed score voting strategy in Appendix \iftoggle{cvprfinal}{\ref{app:iou_voting}}{\iftoggle{cvprrebuttal}{}{F}}.

\section{Implementation Details}
\label{sec:apprachdetails}

In this section, we showcase several implementation details of our approach. 
We include more detailed comments and the overall tracking algorithm in Appendix \iftoggle{cvprfinal}{\ref{app:implementationdetails}}{\iftoggle{cvprrebuttal}{}{A}}.

\noindent\textbf{Online Approaches.} Following \cite{DiMP}, we sample training data in a sequence-wise scheme with a support set $Z$ and a query set $Z'$, each with $3$ and $2$ frames. 
The inner-loop of online approaches, including ONR and ONC is performed with $Z$ under discriminative learning loss for fast convergence, where target mask and spatial weight are data-driven. The steepest gradient descent (SGD) with Gauss-Newton simplification is performed. 
For regression, the weight of the label generator (the coefficients of a radial basis function) is initialized as Gaussian and end-to-end trained with the network. 
For classification, the anchor boxes having IoU with predicted boxes higher than $0.8$ and the one with the highest IoU are assigned as $1$ while those lower than $0.3$ as $0$. The left ones are set to $-1$ and ignored. 
For the outer-loop, the model is trained with $Z'$ under hinge loss to avoid the dominance of easy negatives.
The exact online learning formulation, loss function, and online tracking algorithm can be found in Appendix.

\noindent\textbf{Offline Approaches.} Following \cite{DaSiamRPN,SiamRPN++}, we conduct a pair-wise training scheme with negatives from randomly-sampled other videos\footnote{If the offline and online approaches are jointly trained, we use the sequence-wise scheme instead. We take the averaged feature of $Z$ as template and exclude negative pairs.}. 
While the training loss for classification is set to recently-proposed focal loss \cite{RetinaNet} to tackle class imbalance, the loss for training regression is a penalty-reduced variant of focal loss \cite{CenterNet} to account for continuous values. The loss of negatives and positives samples are split and normalized with each of their total amount such that background and foreground contribute equally to the total loss.
The training label subject to Gaussian distribution for offline regression is centered at the target center with a variance set as a hyper-parameter according to the search window size, while the training label subject to Bernoulli distribution for offline classification is assigned using a modified version of recently-proposed ATSS \cite{ATSS}. More details can be found in Appendix.

\noindent\textbf{Network Architecture.} We use the original ResNet with interpolation instead of dilated ResNet for feature extraction (see Appendix \iftoggle{cvprfinal}{\ref{app:interpolation}}{\iftoggle{cvprrebuttal}{}{C}} for details). Three $1\times1$ convolution layers are attached to $block2-4$ output reducing the dimension to $256$. 
For offline approaches, an extra head is attached after depth-wise correlation reducing dimension from $256$ to $A$ (5 for OFC-5$s$, $1$ for OFC-1$s$ and OFR). The results are weighted sum over $3$ layers. 
For online approaches, we first fuse features via weighted sum and have it up-channel correlated with the online updated weight $\varphi$, outputting the result with dimension $A$ (5 for ONC-5$s$, $1$ for ONC-1$s$ and ONR). Figure \ref{fig:architecture} shows the overall architecture with OFC for accurate localization $s^a_\theta$ and ONR for robust localization $s^r_\theta$. 
For target representation, the derivation of bounding boxes $b_\theta$ resembles the offline approaches for target localization. A box head is attached for each of $3$ layers after depth-wise correlation with the template reducing dimension from $256$ to $4\times A$. The IoU $o_\theta$ for score voting is obtained by adding an extra slot to the box head instead of introducing an extra branch, yielding $(\triangle cx,\triangle cy,\triangle w,\triangle h, o)$ for each anchor scale and overall output dimension being $5\times A$. Note that for the dual-modal localization framework, the model's output with smaller $A$ is broadcast to the size of the larger one to obtain the representation result. 

\noindent\textbf{Training Setup.} 
The training strategy is the same as \cite{SiamRPN++}.
The datasets we used for training include: ImageNet VID \cite{imagenet}, Youtube-BB  \cite{youtubebb}, COCO \cite{coco}, LaSOT \cite{lasot}, and GOT-10k \cite{got10k}. Our method is implemented in Python with PyTorch, and the experiments are conducted on a computer with NVIDA GeForce GTX 1080ti GPUs. Following the discussions about loss of ONR and OFC above, together with losses for box offsets and IoU prediction, the total loss can derived as
\begin{equation}
\begin{aligned}
\label{form:loss}
    L_{\mathrm{total}} = &\lambda_rL_r(s^r_\theta,y^r) + \lambda_aL_a(s^a_\theta,y^a) \\
    &+ \lambda_bL_b(b_\theta, y^b) + \lambda_o L_o(o_\theta, y^o)
\end{aligned}
\end{equation}
where $\lambda_r$, $\lambda_a$, $\lambda_b$, and $\lambda_o$ are re-weighting parameters and are set to $1$, $10$, $1.2$, and $1.2$ respectively. $L_r$, $L_a$, $L_b$, and $L_o$ are focal loss, hinge loss, L1 loss, and binary cross-entropy (BCE) loss respectively. $y^r$, $y^a$, $y^b$, and $y^o$ are training labels consisting of Gaussian heatmap, Bernoulli heatmap, box offsets, and box IoU respectively. 

\section{Experiments}

In this section, we first investigate the effectiveness of the proposed dual-modal localization framework and IoU voting strategy. Further, we evaluate our method on $8$ benchmarks and compare the results with previous trackers. 

\subsection{Ablation Study}

We perform extensive ablation studies to showcase the validity of our proposed method. The experiments are conducted on the dataset of VOT2018 and NUO323 (a combination of NFS100 (30fps), UAV123, and OTB100).

\begin{table}[!t]
\setlength{\tabcolsep}{2.2mm}{
\centering
\begin{tabu}{c|c|c|c|c|c} 
\multirow{2}{*}{formulation}  & \multicolumn{3}{c|}{VOT2018} & \multicolumn{2}{c}{NUO323} \\  
\cline{2-6}
& A$\uparrow$ & R$\downarrow$ & EAO$\uparrow$ & AUC$\uparrow$ & NPr$\uparrow$ \\
\Xhline{1.2pt} 
 OFR & 0.586 & 0.262 & 0.386 & 0.599 & 0.768 \\
 OFC-$1s$ & 0.601 & 0.225 & 0.394 & 0.601 & 0.769 \\
 OFC-$5s$ & 0.598 & 0.192 & 0.432 & 0.609 & 0.777 \\
 ONR & 0.608 & 0.108 & 0.511 & 0.651 & 0.833 \\
 ONC-$1s$ & 0.578 & 0.089 & 0.488 & 0.640 & 0.814 \\
 ONC-$5s$ & 0.587 & 0.108 & 0.496 & 0.646 & 0.821 \\
\hline
 Ours & \color{red}{\bf{0.612}} & \color{red}{\bf{0.084}} & \color{red}{\bf{0.540}} & \color{red}{\bf{0.665}} & \color{red}{\bf{0.844}} \\
\end{tabu}}
\vspace{0.05cm}
\caption{Uni-Modal vs. Dual-Modal. The dual-modal localization framework with OFC-5$s$ and ONR (bottom) achieves the top-performing results.} 
\label{tab:dualmodal}
\vspace{-0.3cm}
\end{table}

\noindent\textbf{Uni-Modal vs. Dual-Modal.} To illustrate the effectiveness of dual-modal architecture, we compare our method with the other uni-modal localization methods. For a fair comparison, all the other components (i.e., box branch, feature extraction, etc.) besides the localization part are kept identical.
As illustrated in Table \ref{tab:dualmodal}, online approaches generally demonstrate a stronger ability in terms of robustness according to R of VOT2018 dataset, while offline approaches have better accuracy with higher A. Among these formulations, ONR shows the best performance which indicates that robust localization is of overriding significance for good tracking. 
Our methods built upon ONR and OFC-$5s$ reaches an EAO $0.540$ on VOT2018 and an AUC $0.665$ on NUO323, outperforming all $4$ uni-modal localization methods by a large margin.

\begin{table}[!t]
\setlength{\tabcolsep}{1.4mm}{
\centering
\begin{tabu}{c|c|c|c|c|c|c}
\multirow{2}{*}{offline} & \multirow{2}{*}{online}  & \multicolumn{3}{c|}{VOT2018} & \multicolumn{2}{c}{NUO323} \\
\cline{3-7}
& & A$\uparrow$ & R$\downarrow$ & EAO$\uparrow$ & AUC$\uparrow$ & NPr$\uparrow$ \\
\Xhline{1.2pt}
OFR & \multirow{3}{*}{ONC-$5s$} & 0.604 & 0.122 & 0.512 & 0.648 & 0.823 \\
OFC-$1s$ & & 0.598 & 0.112 & 0.520 & 0.650 & 0.825 \\
OFC-$5s$ & & 0.601 & 0.098 & 0.533 & 0.656 & 0.837 \\
\hline
OFR & \multirow{3}{*}{ONC-$1s$} & 0.587 & 0.098 & 0.497 & 0.644 & 0.822 \\
OFC-$1s$ & & 0.596 & 0.112 & 0.498 & 0.648 & 0.829 \\
OFC-$5s$ & & 0.598 & 0.089 & 0.514 & 0.650 & 0.828 \\
\hline
OFR & \multirow{3}{*}{ONR} & 0.608 & 0.112 & 0.527 & 0.653 & 0.836 \\
OFC-$1s$ & & 0.592 & 0.112 & 0.523 & 0.656 & 0.840 \\
OFC-$5s$ & & \color{red}{\bf{0.612}} & \color{red}{\bf{0.084}}  & \color{red}{\bf{0.540}}  & \color{red}{\bf{0.665}}  & \color{red}{\bf{0.844}}  \\
\end{tabu}}
\vspace{0.05cm}
\caption{Ablation study on other (offline + online) combinations for dual-modal framework.}
\vspace{-0.2cm}
\label{tab:crossover}
\end{table}

\begin{figure}[!t]
\centering
\includegraphics[width=8.2cm]{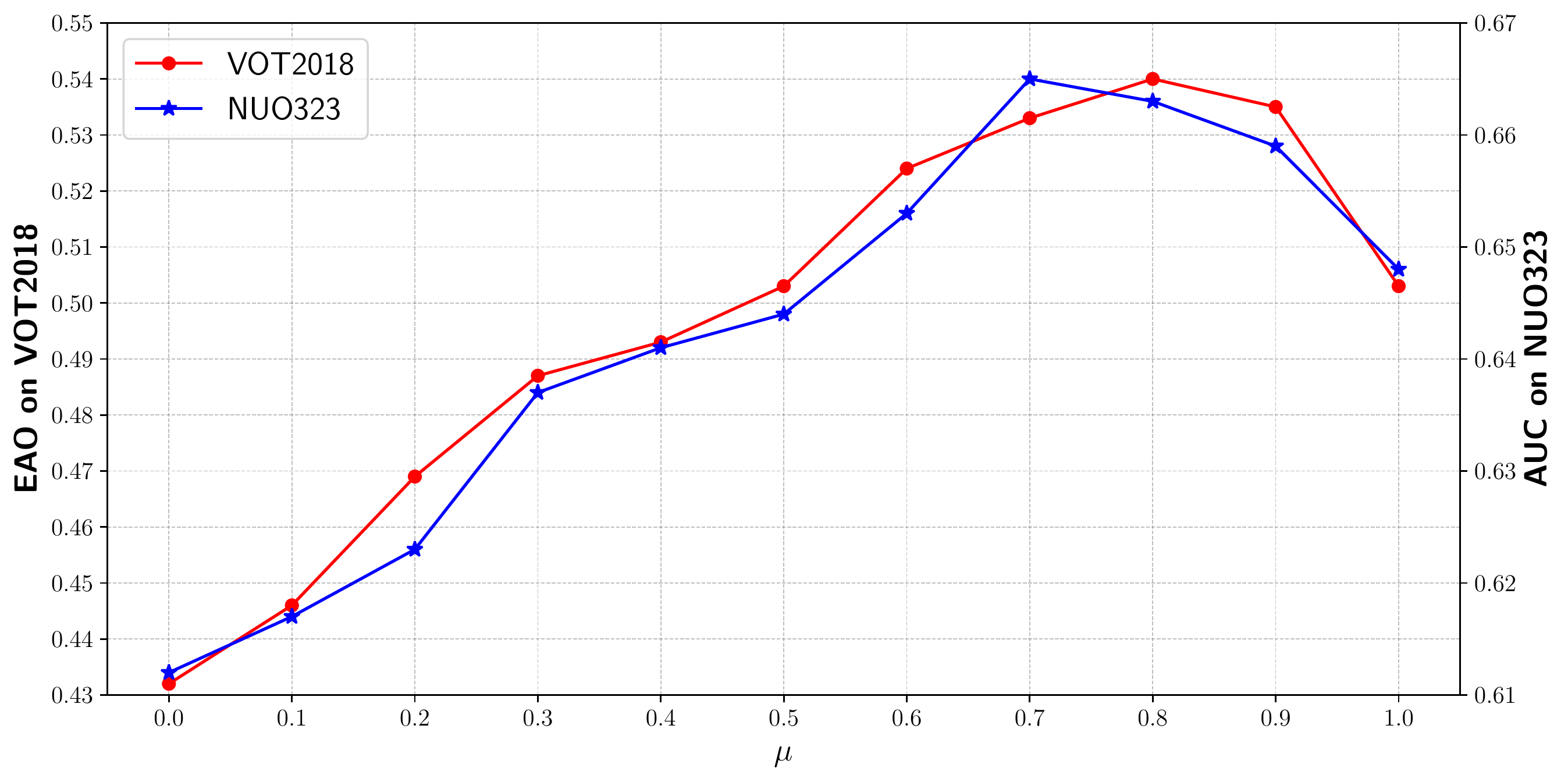}
\caption{EAO on VOT2018 and AUC on NUO323 with varying fusion weight $\mu$.}
\label{fig:varying_mu}
\vspace{-0.5cm}
\end{figure}

\begin{table}[!t]
\setlength{\tabcolsep}{2.1mm}{
\centering
\begin{tabu}{c|c|c|c|c|c} 
\multirow{2}{*}{score voting}  & \multicolumn{3}{c|}{VOT2018} & \multicolumn{2}{c}{NUO323} \\  
\cline{2-6}
& A$\uparrow$ & R$\downarrow$ & EAO$\uparrow$ & AUC$\uparrow$ & NPr$\uparrow$ \\
\Xhline{1.2pt} 
 w/o & \color{red}{\bf{0.612}} & 0.084 & 0.540 & 0.665 & 0.844 \\
 w/ & 0.608 & \color{red}{\bf{0.080}} & \color{red}{\bf{0.564}} & \color{red}{\bf{0.671}} & \color{red}{\bf{0.859}} \\
\end{tabu}}
\vspace{0.05cm}
\caption{Ablation study on the effectiveness of proposed score voting strategy for final box representation. w/ and w/o denotes voting strategy is and is not applied respectively. } 
\label{tab:iouvoting}
\vspace{-0.5cm}
\end{table} 

\begin{figure*}[!t]
\centering
\includegraphics[width=17.5cm]{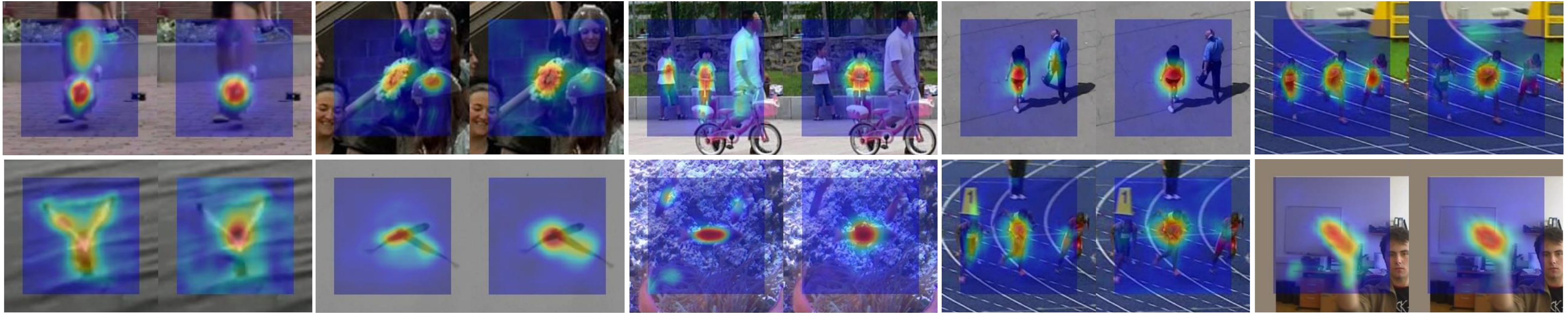}
\caption{Visualization of the heatmap from two localization models. The left figure of each pair is heatmap of accurate localization, while the right one is the heatmap of robust localization. The output is maximized long the extra box scale dimension if with that.}
\label{fig:visualization}
\end{figure*}

\noindent\textbf{Exhausting Dual-Modal Combinations. } As shown in Table \ref{tab:dualmodal}, while online approaches excel at robust localization, the offline approaches are good at accurate localization. Thus we further conduct a cross-over analysis by assigning different labels to the dual-modal framework, which consists of an online branch for robust localization and an offline branch localization. It can be observed from Table \ref{tab:crossover} that, online branch with a Gaussian label formulated as ONR generally yield high tracking performance. The combination of OFC-$5s$ and ONC as accurate and robust localization respectively achieve a comparable result as well by achieving $0.642$ of AUC on NUO323. Moreover, the offline branch with Bernoulli-$5s$ as label tends to have higher accuracy than both Bernoulli-$1s$ and Gaussian label, which validates the superiority of OFC-$5s$ in accurate localization. From this above, our proposed dual-modal localization consisting of ONR and OFC is an optimal choice and outperform other combinations.

\noindent\textbf{Varying Fusion Weight.} With the obtained dual-modal framework consisting of ONR and OFC-$5s$, we investigate the effect of varying fusion weight $\mu$ in terms of EAO on VOT2018 and AUC on NUO323. Results are shown in Figure \ref{fig:varying_mu}. We observe that the results are optimal around $0.7 \sim 0.8$. The results also indicate that robust localization is more crucial for good tracking performance. We set $\mu$ to $0.8$ by default.

\noindent\textbf{Score Voting Strategy.} Next, we further illustrate the effectiveness of our proposed score voting strategy. From Table \ref{tab:iouvoting}, with the introduction of score voting, the EAO of VOT2018 further rises from $0.54$ to $0.564$ and the AUC of NUO323 rises from $0.665$ to $0.671$.
The consistent performance gain over two major benchmarks thus verifies the effectiveness of our proposed score voting strategy to account for uncertainty in the box representation.

\noindent\textbf{Visualization of Heatmaps. } To provide a demonstrative illustration of the superiority of each localization model, we showcase the heatmaps from the dual-modal framework over various sequences. As shown in Figure \ref{fig:visualization}, the first row illustrates the superiority of robust localization with heatmaps from robust localization attending only to the target while those from accurate localization attending to other similar areas. In the scenarios of the presence of distractors, the robust localization can locate a rough region of the target more discriminatively thus greatly reduce the possibility of target drift. The second row of the figures illustrates the superiority of accurate localization over the robust one. While robust localization roughly locating the target, accurate localization is more customary to the target's size thus more precise in locating the optimal point to further generate bounding boxes. Though other peaks occurred, the peak shape of accurate localization around the target is more dependent on the target's original form. From these figures, the effectiveness of the framework of proposed dual-modal localization is verified given the complementary characteristics of two localization models.

\subsection{Comparison with State-of-the-Arts}
We compare our proposed tracker, termed as \textbf{DuML}, with the state-of-the-art approaches on the major benchmarks. Detailed results can be found in Appendix \iftoggle{cvprfinal}{\ref{app:detailed_results}}{\iftoggle{cvprrebuttal}{}{E}}.

\begin{table}[!t]
\footnotesize
\setlength{\tabcolsep}{0.3mm}{
\centering
\begin{tabu}{c|c|c|c|c|c|c|c|c|c} 
 & \tabincell{c}{UPDT \\ \cite{UPDT}} & \tabincell{c}{SiamRPN \\ ++\cite{SiamRPN++}} & \tabincell{c}{ATOM \\ \cite{ATOM}} & \tabincell{c}{DiMP \\ \cite{DiMP}}  & \tabincell{c}{DROL \\ \cite{DROL}} & \tabincell{c}{Ocean \\ \cite{Ocean}} & \tabincell{c}{FCOT \\ \cite{FCOT}} & \tabincell{c}{RPT \\ \cite{RPT}} & Ours \\
\Xhline{0.8pt}
A$\uparrow$ & 0.536 & 0.600 & 0.590 & 0.597 & \green{0.616} & 0.592 & 0.600 & \red{0.629} & \blue{0.608} \\
R$\downarrow$ & 0.184 & 0.234 & 0.204 & 0.153 & 0.122 & 0.117 & \blue{0.108} & \green{0.103} & \red{0.080} \\
EAO$\uparrow$ & 0.378 & 0.414 & 0.401 & 0.440 & 0.481 & 0.489 & \blue{0.508} & \green{0.510} & \red{0.564} \\
\end{tabu}}
\vspace{0.05cm}
\caption{Results on VOT2018 challenge dataset \cite{vot2018} in terms of expected average overlap (EAO), robustness (R), and accuracy (A). \red{Red}, \green{green} and \blue{blue} denote top-$3$ results.} 
\label{tab:vot2018}
\vspace{-0.2cm}
\end{table} 

\begin{table}[!t]
\footnotesize
\setlength{\tabcolsep}{0.22mm}{
\centering
\begin{tabu}{c|c|c|c|c|c|c|c|c|c} 
 & \tabincell{c}{SiamM- \\ ask\cite{SiamMask}} & \tabincell{c}{SiamRPN \\ ++\cite{SiamRPN++}} & \tabincell{c}{ATOM \\ \cite{ATOM}} & \tabincell{c}{STN \\ \cite{STN}}  & \tabincell{c}{Ocean \\ \cite{Ocean}} & \tabincell{c}{DiMP \\ \cite{DiMP}} & \tabincell{c}{DRNet \\ \cite{vot2019}} & \tabincell{c}{RPT \\ \cite{RPT}} & Ours \\
\Xhline{0.8pt}
A$\uparrow$ & 0.594 & 0.580 & \blue{0.603} & 0.589 & 0.594 & 0.594 & \green{0.605} & \red{0.623} & \green{0.605} \\
R$\downarrow$  & 0.461 & 0.446 & 0.411 & 0.349 & 0.316 & 0.278 & \blue{0.261} & \green{0.186} & \red{0.176} \\
EAO$\uparrow$ & 0.287 & 0.292 & 0.301 & 0.314 & 0.350 & 0.379 & \blue{0.395} & \green{0.417} & \red{0.437} \\
\end{tabu}}
\vspace{0.05cm}
\caption{Results on VOT2019 challenge dataset \cite{vot2019} in terms of EAO, R, and A.} 
\label{tab:vot2019}
\vspace{-0.2cm}
\end{table} 

\noindent\textbf{VOT2018 \cite{vot2018}.} VOT2018 dataset consists of $60$ challenging videos. 
The tracker's overall performance is evaluated upon robustness and accuracy, defined using failure rate and IoU, and a comprehensive protocol EAO involves both two respects. We compare our methods with the state-of-the-art methods as shown in Table \ref{tab:vot2018}, resulting in an absolute gain on EAO from $0.510$ to $0.564$ with a $10.6\%$ relative gain. Moreover, our tracker achieves the best robustness among all existing methods while maintaining a desirable accuracy.

\noindent\textbf{VOT2019 \cite{vot2019}.} Compared with VOT2018, VOT2019 dataset includes more challenging sequences for evaluating the tracker's robustness and accuracy. As shown in Table \ref{tab:vot2019}, our tracker consistently achieves a state-of-the-art result by outperforming the previously top tracker by $4.6\%$ and resulting in an EAO of $0.437$ and robustness of $0.176$.

\begin{table}[!t]
\footnotesize
\setlength{\tabcolsep}{0.08mm}{
\centering
\begin{tabu}{c|c|c|c|c|c|c|c|c|c} 
 & \tabincell{c}{MDNet \\ \cite{MDNet}} & \tabincell{c}{ECO \\ \cite{ECO}} & \tabincell{c}{SiamRPN \\ ++\cite{SiamRPN++}} & \tabincell{c}{ATOM \\ \cite{ATOM}}  & \tabincell{c}{DiMP \\ \cite{DiMP}} & \tabincell{c}{PrDiMP \\ \cite{PrDiMP}} & \tabincell{c}{DROL \\ \cite{DROL}} & \tabincell{c}{RPT \\ \cite{RPT}} & Ours \\
\Xhline{0.8pt}
OTB2015 & 67.8 & 69.1 & \blue{69.6} & 66.9 & 68.4 & 69.5 & \green{71.5} & \green{71.5} & \red{71.6} \\
NFS & 42.2 & 46.6 & - & 58.4 & \blue{62.0}& \red{63.8} & - & - & \green{63.6} \\
\end{tabu}}
\vspace{0.05cm}
\caption{Results on OTB2015 and NFS (30fps) datasets \cite{otb2015,nfs} in terms of overall AUC score. } 
\label{tab:otbnfs}
\vspace{-0.5cm}
\end{table} 

\noindent\textbf{OTB2015 \cite{otb2015}.} OTB2015 dataset contains a total amount of $100$ sequences with motion, scale change, and illumination change, etc. The evaluation protocol is over precision plot and success plot (AUC). As shown in Table \ref{tab:otbnfs}, we achieve a state-of-the-art performance with an AUC of $71.6$ and a precision of $93.7$. 

\noindent\textbf{NFS (30fps) \cite{nfs}.} The $30$ fps version of dataset Need for Speed (NFS) resembles the evaluation protocol of OTB2015. With the fast motion of targets, along with challenging scenarios like distractors and scale change, NFS functions well for a comprehensive evaluation benchmark. As shown in Table \ref{tab:otbnfs}, our tracker achieves an AUC score of $63.6$ and a precision score of $77.4$, indicating a desirable tracking capability.

\begin{figure}[!t]
\centering
\subfigure{\includegraphics[width=0.49\linewidth,height=0.49\linewidth]{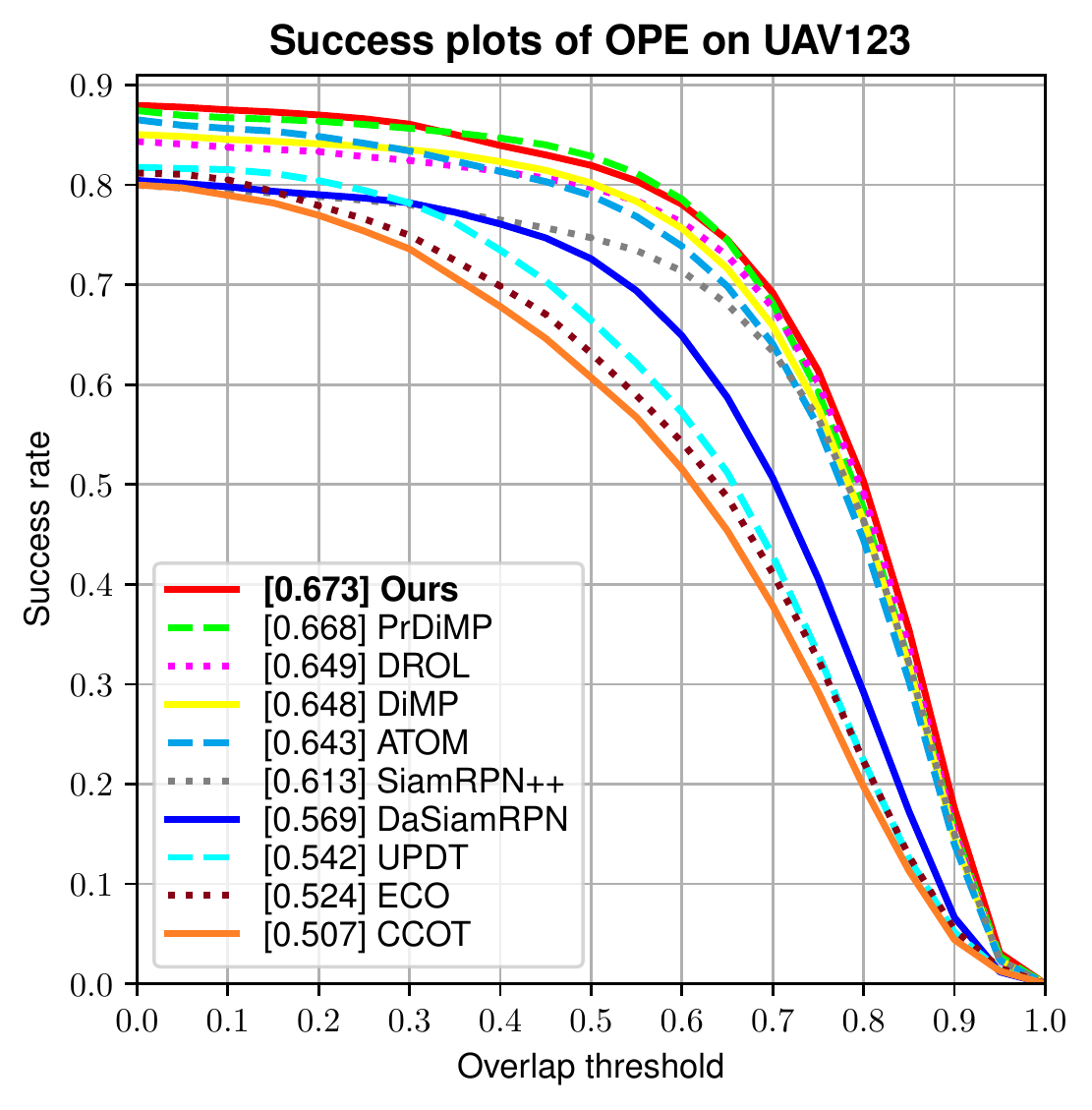}}
\subfigure{\includegraphics[width=0.49\linewidth,height=0.49\linewidth]{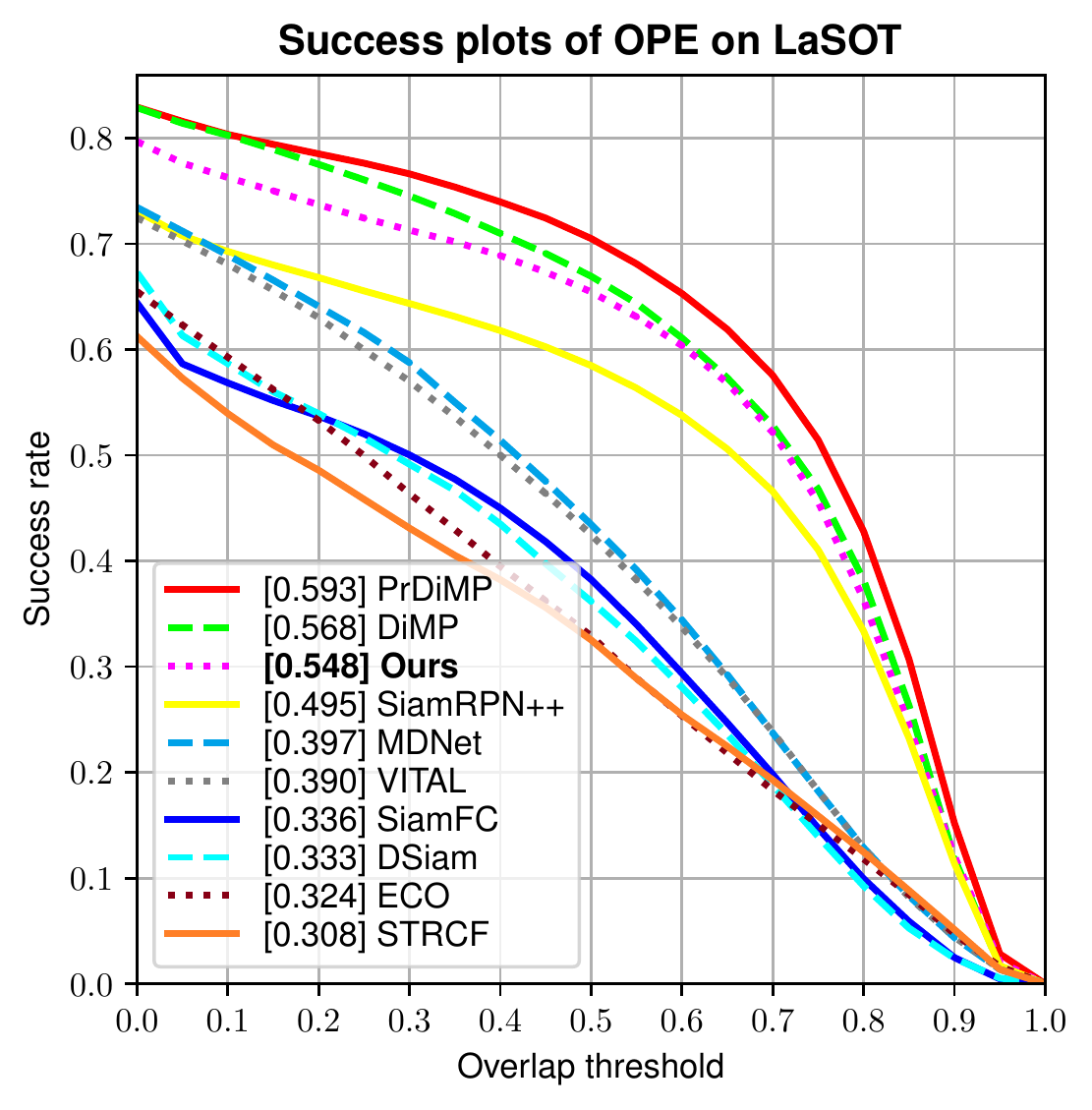}}
\vspace{-0.05cm}
\caption{Results in terms of overall AUC score on UAV123 dataset \cite{uav123} (left) and LaSOT dataset (right).}
\vspace{-0.2cm}
\label{fig:uav_and_lasot}
\end{figure}

\noindent\textbf{UAV123 \cite{uav123}.} UAV123 dataset contains $123$ sequences collected from a UAV perspective.
, which are practical in reality considering one major application of tracking algorithm is for UAVs' purpose. 
As shown in Table \ref{fig:uav_and_lasot}, our tracker achieved an AUC score of $67.3$ and a precision score of $88.4$, with a large performance gain compared with the previous state-of-the-art trackers.

\noindent\textbf{LaSOT \cite{lasot}} LaSOT is a challenging long-term tracking benchmark consisting of $280$ sequences. 
The target can be out-of-frame and then reappear during tracking. 
From Table \ref{fig:uav_and_lasot}, our proposed method achieve comparable results compared with the top-performing methods, with an AUC score of $54.8$ and a precision score of $56.3$.

\noindent\textbf{TrackingNet \cite{trackingnet}.} TrackingNet is a large-scale tracking benchmark, which consists of over $30k$ sequences in total and $511$ sequences for testing without publicly available groundtruth. The results, shown in Table \ref{tab:trackingnet}, demonstrates that our proposed method achieves state-of-the-art performance in terms of NPr precision and comparable performance in both AUC and precision.

\begin{table}[!t]
\footnotesize
\setlength{\tabcolsep}{0.2mm}{
\centering
\begin{tabu}{c|c|c|c|c|c|c|c|c|c} 
 & \tabincell{c}{SiamFC \\ \cite{SiamFC}} & \tabincell{c}{UPDT \\ \cite{UPDT}} & \tabincell{c}{ATOM \\ \cite{ATOM}} & \tabincell{c}{SPM \\ \cite{SPM}}  & \tabincell{c}{SiamRPN \\ ++\cite{SiamRPN++}} & \tabincell{c}{DiMP \\ \cite{DiMP}} & \tabincell{c}{DROL \\ \cite{DROL}} & \tabincell{c}{SiamA- \\ ttn\cite{SiamAttn}} &  Ours \\
\Xhline{0.8pt}
AUC$\uparrow$ & 57.1 & 61.1 & 70.3 & 71.2 & 73.3 & 74.0 & \blue{74.6} & \red{75.2} & \green{74.7} \\
NPr$\uparrow$  & 66.6 & 70.2 & 77.1 & 77.8 & 80.0 & \blue{80.1} & \green{81.7} & \green{81.7} & \red{81.8} \\
Pr$\uparrow$ & 53.3 & 55.7 & 64.8 & 66.8 & 69.4 & 68.7 & \blue{70.8} & \red{71.5} & \green{70.9} \\
\end{tabu}}
\vspace{0.05cm}
\caption{Results on TrackingNet test set \cite{trackingnet} in terms of precision (Pr), normalized precision (NPr), and success (AUC). } 
\label{tab:trackingnet}
\vspace{-0.2cm}
\end{table} 

\begin{table}[!t]
\footnotesize
\setlength{\tabcolsep}{0.5mm}{
\centering
\begin{tabu}{c|c|c|c|c|c|c|c|c} 
 & \tabincell{c}{SiamFC \\ \cite{SiamFC}} & \tabincell{c}{SiamRPN \\ ++\cite{SiamRPN++}} & \tabincell{c}{ATOM \\ \cite{ATOM}} & \tabincell{c}{DMV \\ \cite{DMV}}  & \tabincell{c}{DiMP \\ \cite{DiMP}} & \tabincell{c}{Ocean \\ \cite{Ocean}} & \tabincell{c}{FCOT \\ \cite{FCOT}} & Ours \\
\Xhline{0.8pt}
$\mathrm{SR}_{0.50}\uparrow$ & 35.3 & 61.8 & 63.4 & 69.5 & 71.7 & \blue{72.1} & \red{76.3} & \green{72.8} \\
$\mathrm{SR}_{0.75}\uparrow$ & 9.8 & 32.9 & 40.2 & \blue{49.2} & \blue{49.2} & 48.7 & \red{51.7} & \green{50.1} \\
AO$\uparrow$ & 34.8 & 51.8 & 55.6 & 60.1 & \blue{61.1} & \blue{61.1} & \red{64.0} & \green{62.3} \\
\end{tabu}}
\vspace{0.05cm}
\caption{Results on GOT-10k test set \cite{got10k} in terms of average overlap (AO), $\mathrm{SR}_{0.75}$, and $\mathrm{SR}_{0.5}$.} 
\label{tab:got10k}
\vspace{-0.5cm}
\end{table} 

\noindent\textbf{GOT-10k \cite{got10k}.} GOT-10k is a recently-proposed large-scale dataset for both training and testing, with no overlap in object classes between training and testing. For GOT-10k test, we train our tracker by only using the GOT10k train split following its standard protocol. From Table \ref{tab:got10k}, our proposed method achieves comparable results with the top methods with an AO of $62.7$ and a $\mathrm{SR}_{0.50}$ of $73.5$.

\section{Conclusion}

In this work, we conduct a systematic investigation into accuracy and robustness for major $4$ formulations existing in the tracking community. To account for the conflicting performance under accuracy and robustness, we analyze the underlying restrictions of each one;s design. 
Further, we resort to the idea of ensemble and propose a dual-modal localization framework to account for separate needs for both robustness and accuracy, with the former solved by ONR and the latter by OFC. 
To account for the uncertainty in target representation, we propose a simple but effective score voting strategy to involve adjacent predicted boxes around the target center.
The effectiveness of our method is validated among the major benchmarks with a large performance gain and a speed beyond the real-time requirement.

{\small
\bibliographystyle{ieee_fullname}
\bibliography{egbib}
}


\appendix
\newpage

\section{Implementation Details}
\label{app:implementationdetails}

\noindent\textbf{Online Approaches (ONC \& ONR) with Discriminative Learning Loss and Steepest Gradient Descent \cite{DiMP}}. We follow the same formulation and update strategy for online learning as DiMP \cite{DiMP}. Specifically, the inner-loop is trained under the least-square error with the residule defined as 
\begin{equation}
r(p, c)= v_c \cdot (m_cp + (1-m_c)\ \mathrm{max}(0,p)-y_c)
\end{equation}
where p is the prediction and c is the target center. $v_c$, $m_c$, and $y_c$ are offline trained as the coefficients of a set of triangular basis functions. Further, we use steepest gradient descent, which scales the step length with its Hessian matrix $Q$ and further the inner product of Jacobian vector as $Q=J^TJ$, formulated as
\begin{equation}
\label{eq:rr-prim-itr-update}
    \theta^{(i+1)} = \theta^{(i)} - \frac{\triangledown L(\theta^{(i)})^T\triangledown L(\theta^{(i)})}{\triangledown L(\theta^{(i)})^TQ\triangledown L(\theta^{(i)})} \triangledown L(\theta^{(i)})
\end{equation}
where $i$ denotes the number of iterations of stepwise optimization. We initialize the parameter $\theta^{(0)}$ with template feature adaptively pooed (instead of using precise RoI pooling \cite{PrecisePool}) to the kernel size $5 \times 5$.

\begin{table}[b]
\setlength{\tabcolsep}{1.5mm}{
\centering
\begin{tabu}{c|c|c|c|c|c} 
\multirow{2}{*}{form} & label & \multirow{2}{*}{loss} & \multicolumn{3}{c}{VOT2018} \\  
\cline{4-6}
& assignment & & Acc & R & EAO \\
\Xhline{1.2pt} 
\multirow{4}{*}{OFC-5$s$} & IoU & CE & 0.601 & 0.201 & 0.414  \\
& IoU & FC & 0.608 & 0.215 & 0.419  \\
& ATSS$_{Min\_L2}$ & FC & 0.596 & 0.201 & 0.425 \\
& ATSS$_{Max\_IoU}$ & FC & 0.598 & 0.192 & 0.432 \\
\hline
\multirow{2}{*}{OFC-1$s$} & ATSS$_{Min\_L2}$ & FC & 0.604 & 0.231 & 0.386 \\
& ATSS$_{Max\_IoU}$ & FC & 0.601 & 0.225 & 0.394 \\
\hline
\multirow{2}{*}{OFR} & Gaussian & FC$_{RG}$ & 0.594 & 0.276 & 0.375 \\
& Gaussian & FC$_{PR}$ & 0.586 & 0.262 & 0.386
\end{tabu}}
\vspace{0.05cm}
\caption{Ablation study on different modifications for OFC and OFR. } 
\label{tab:offline}
\vspace{-0.5cm}
\end{table} 

\noindent\textbf{Offline Classification (OFC) with Focal Loss \cite{RetinaNet} and ATSS \cite{ATSS}}. To further alleviate the problem of class imbalance, we utilized focal loss for training OFC. As for label assignment, \cite{SiamRPN++} assigns the positive and negative according to the computed IoU between anchor boxes and groundtruth. The recently-proposed ATSS \cite{ATSS} assigns the positive and negative according to the pre-computed statistics which bridges the gap between anchor-based and anchor-free detectors. The original ATSS computes the mean and variance of boxes whose center is closest to the center of groundtruth based on L$2$ distance. We denote it as ATSS$_{Min\_L2}$. In practice, we compute statistics of top-$15$ anchor boxes for OFC-5$s$ and top-$11$ anchor boxes for OFC-1$s$ having the highest IoU with the groundtruth to assign the positive and negative anchors, which yield higher performance. We denote it as ATSS$_{Max\_IoU}$. Our modifications and implementations for OFC serve as a stronger baseline compared to SiamRPN++, shown in Table \ref{tab:offline}.

\noindent\textbf{Offline Regression (OFR) with Continuous Focal Loss}. We consider the continuous variants of focal loss to train the offline regression. Specifically, we consider the penalty-reduced focal loss \cite{CenterNet}, denoted as FC$_{PR}$, where the pixels at the center of Gaussian as positive examples while the others as negatives whose loss is reduced corresponding to its continuous labels, formulated as 
\begin{equation}
L = -\frac{1}{N}\sum_y \begin{cases}
(1-P_y)^\alpha log(P_y) & \text{if Y$_y$ = 1} \\ 
(1-Y_y)^\beta P_y^\alpha log(1-P_y) & \text{otherwise}
\end{cases}
\end{equation}
where $\alpha$ and $\beta$ are hyper-parameters. While the continuous value can be regressed, the local optima of this loss is not the groudtruth label. To tackle this problem, we have tried another similar loss, denoted as regressive focal loss, with $P=Y$ holds true at the optimal point. The FC$_{RG}$ can be formulated as
\begin{equation}
L = -\frac{1}{N}\sum_y  - Y_y^\beta(Y_y-P_y)^\alpha log(P_y)
\end{equation}
where $\alpha \in \{2n \ | \ n\in N^+\}$ and $\beta$ are hyper-parameters. We set $\alpha$ to $2$ and $\beta$ to $4$ for both losses. We observe that FC$_{PR}$ works better in practice, shown in \ref{tab:offline}. We use FC$_{PR}$ in all our experiments. 

\noindent\textbf{Tracking Algorithm.} During tracking, a support set with memory size $M=50$ is maintained for ONR to record historical glances of the target with a fixed space of augmented initial frames. For initialization, the weight is optimized for $10$ iterations with an augmented support set obtained from the first frame. For online tracking, the update is conducted with a learning rate of $0.1$ and $2$ iterations every $20$ frame, or a learning rate of $0.2$ and $1$ iteration once the possible distractors are detected. If the target is lost, we discard that frame. We recommend the readers to refer to \cite{DiMP} for more technical details. After fusing the heatmap of ONR and OFC, we apply the same post-processing process as \cite{SiamRPN} (i.e., cosine window, box temporal smoothing, etc.) We recommend the readers to refer to \cite{SiamRPN} for more details.
The search region is set to $255 \times 255$ by default and $319 \times 319$ for NFS30, UAV123, and LaSOT. The $\epsilon$ in Equation $(8)$ is set to $0.01$. The $\sigma$ in Equation $(9)$ is set to $0.0025$.

\section{Faster Speed with Interpolation.} 
\label{app:interpolation}

The successful application of deep networks \cite{SiamRPN++} like ResNet-50 \cite{ResNet} is crucial on desirable performance of a tracker. Since the search region has generally the same resolution by rescaling the target from the original image, a modified ResNet with the stride of the downsampled convolution from the last two blocks changing from $2$ to $1$. To increase the receptive field, dilated convolutions with a rate $2$ are applied. We instead propose to use the original ResNet. To ensure the same feature size for layerwise aggregation, we interpolate the features from the last two blocks to the same size of feature from block 2 (with stride $8$). Compared with the original approach, the interpolation greatly saves the computation cost from the last two blocks since the input feature has a smaller spatial size due to downsampling. Compared with the baseline of SiamRPN++ operating at $46$ FPS on an NVIDIA Titan Xp GPU, the offline classification module alone in our framework can operate at $90$ FPS, achieving a $96\%$ speed gain. 

\begin{table}[!t]
\setlength{\tabcolsep}{2.1mm}{
\centering
\begin{tabu}{c|c|c|c|c|c} 
\multirow{2}{*}{training data}  & \multicolumn{3}{c|}{VOT2018} & \multicolumn{2}{c}{NUO323} \\  
\cline{2-6}
& Acc & R & EAO & AUC & NPr \\
\Xhline{1.2pt} 
 VYCLG & 0.608 & 0.080 & 0.564 & 0.671 & 0.859 \\
 VYCD & 0.602 & 0.089 & 0.539 & 0.664 &  0.856 \\
 VYC & 0.604 & 0.103 & 0.530 & 0.661 & 0.846 \\
 VC & 0.588 & 0.098 & 0.519 & 0.646 & 0.835
\end{tabu}}
\vspace{0.05cm}
\caption{Ablation study on the effect of training data. } 
\label{tab:trainingdata}
\vspace{-0.5cm}
\end{table}

\label{detailresult}

\begin{figure*}[!t]
\centering
\subfigure{\includegraphics[height=7cm,width=8.5cm]{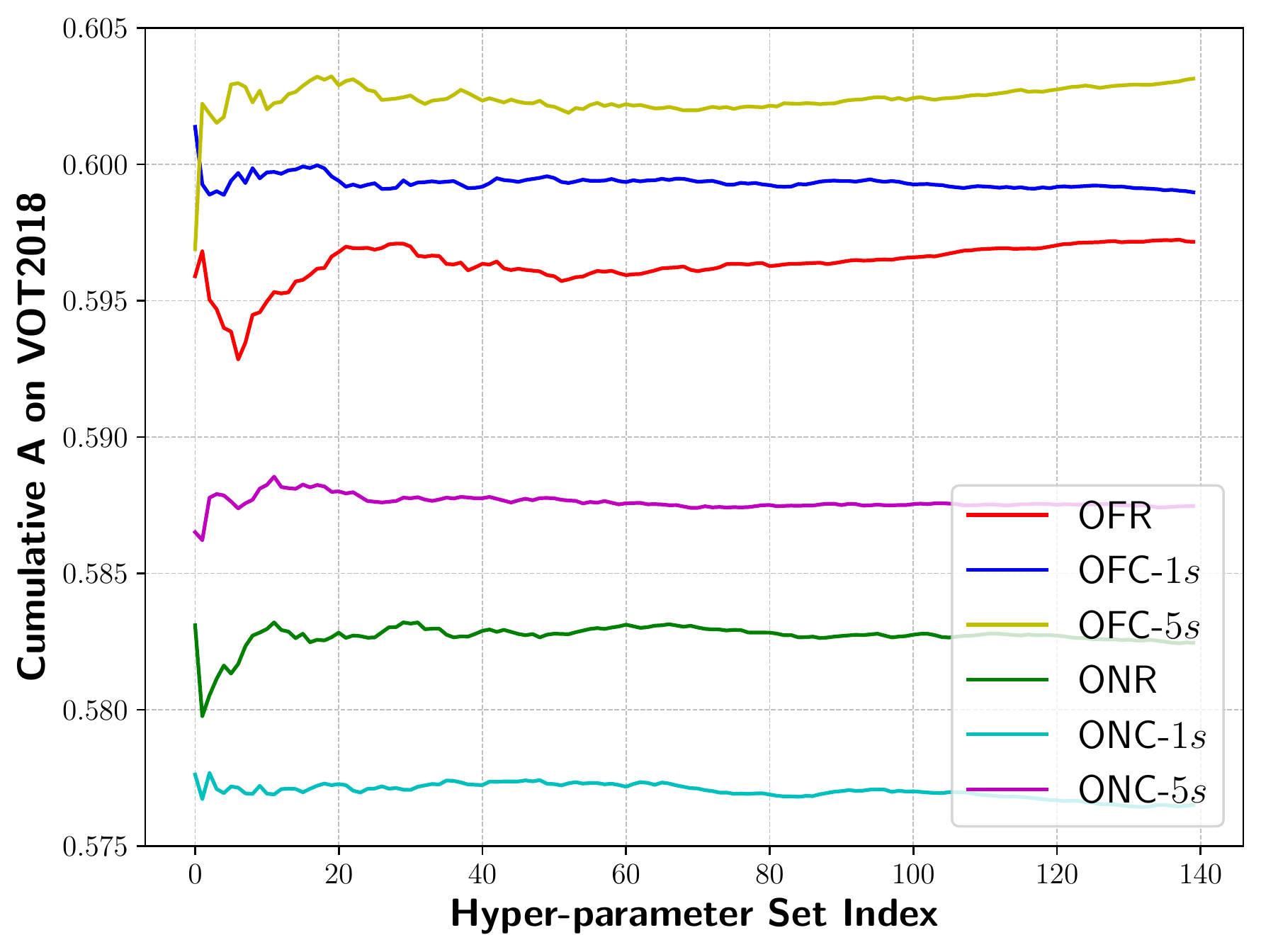}}
\subfigure{\includegraphics[height=7cm,width=8.5cm]{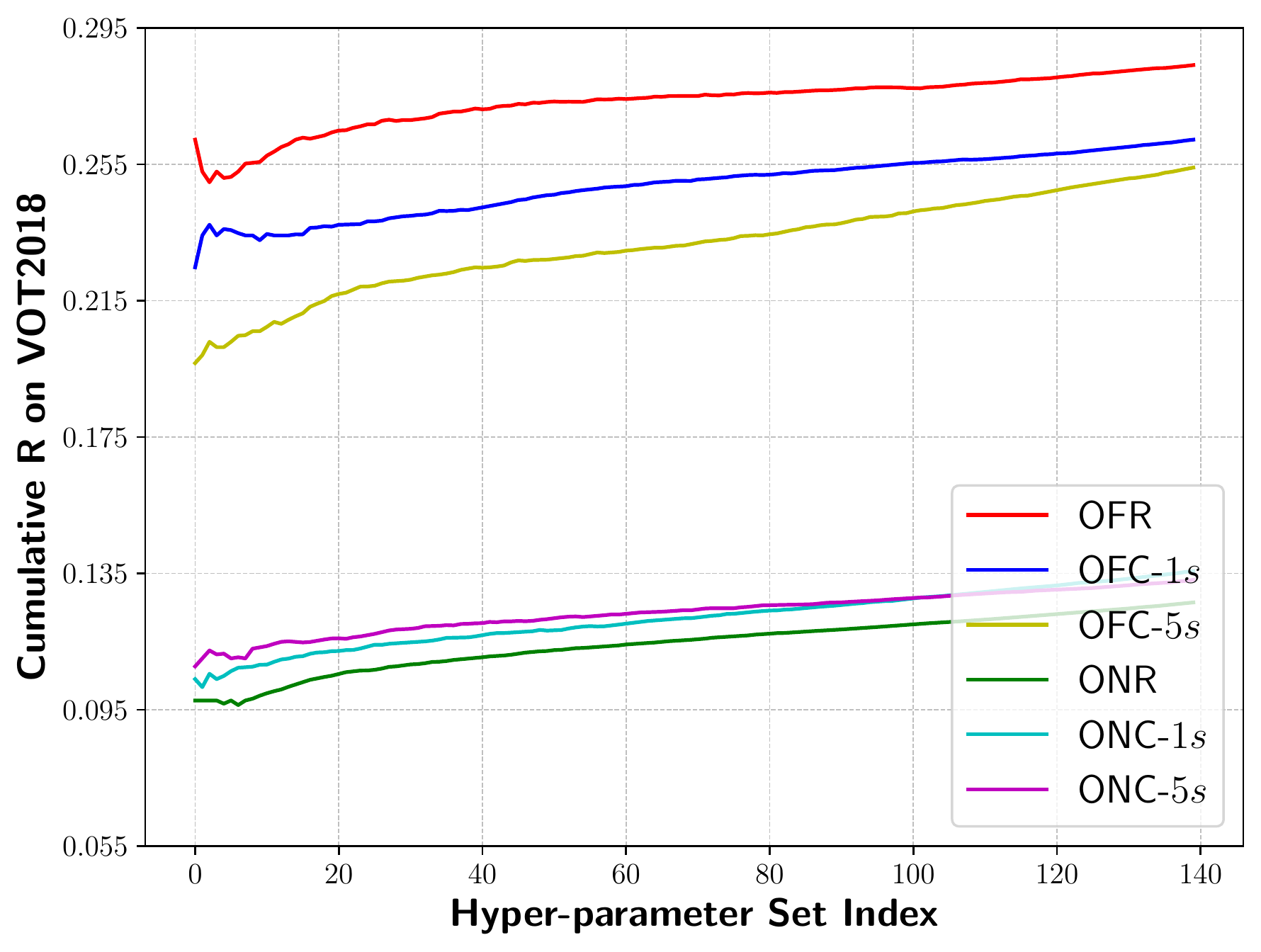}}
\vspace{-0.05cm}
\caption{Cumulative average of Accuracy (left) and Robustness (right) on VOT2018. }
\label{fig:cumulative}
\vspace{-0.5cm}
\end{figure*}

\section{Cumulative A and R on VOT2018}
\label{app:cumulative}

Since we follow the same post-processing process as \cite{SiamRPN}, several hyper-parameters are introduced during tracking. We observe that A and R can vary drastically with different sets of hyper-parameters (i.e., window influence, penalty\_k, etc.). A model with a higher R will often yield a higher A (The accumulation error will be eliminated more often with more re-initialization). Therefore, the comparison can be substantially nullified if on a fixed set of hyper-parameters or their respective optimal set of hyper-parameters. To account for the problem and conduct a more trustworthy comparison, we report the average results of each model over multiple sets of hyper-parameters ($140$ in practice). We report the cumulative mean of A and R on these $140$ runs. As shown in Figure \ref{fig:cumulative}, the averaged A and R are more stable for a stable comparison and analysis.

\section{Impact of Training Data}
Our original tracker is trained using a collection of ImageNet VID (V) \cite{imagenet}, Youtube-BB  (Y) \cite{youtubebb}, COCO (C) \cite{coco}, LaSOT (L) \cite{lasot}, and GOT-10k (G) \cite{got10k}. The original full model is trained using VYCLG. For a more fair comparison with previous works \cite{SiamRPN++}, we train our tracker using VYCD. We also study the models with less data - VYC and VC. We evaluate the performance on VOT2018 and NUO323. Results are reported in Table \ref{tab:trainingdata}. We can see that the performance of the model trained with VYCD only drops a little and still achieves state-of-the-art performance. The model trained with minimal data can still achieve an EAO of $0.520$ on VOT2018 and an AUC of $0.646$ on NUO323, demonstrating the strong generalization ability of our proposed method.

\begin{figure*}[!t]
\centering
\subfigure{\includegraphics[width=0.33\linewidth,height=0.33\linewidth]{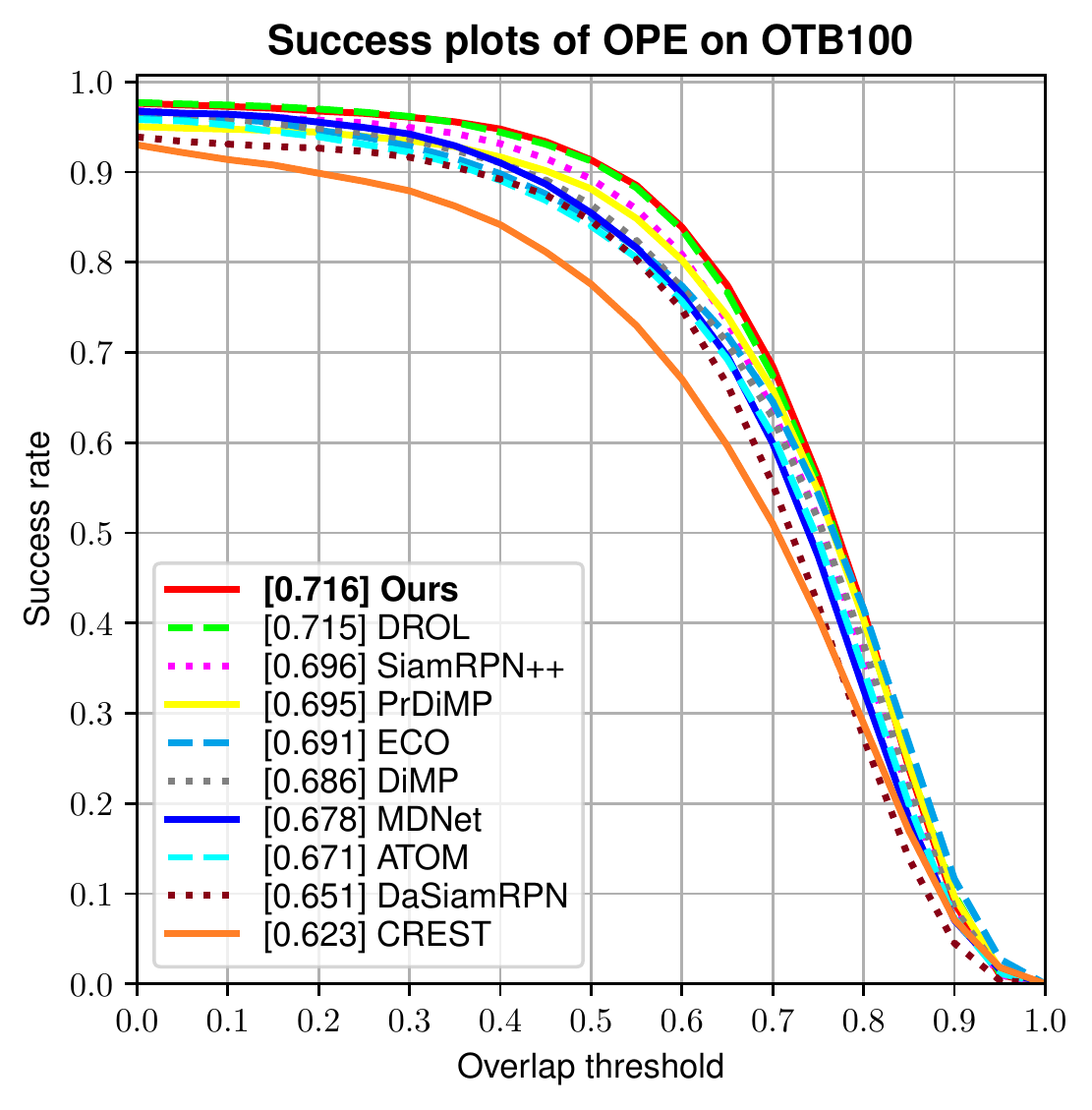}}
\subfigure{\includegraphics[width=0.33\linewidth,height=0.33\linewidth]{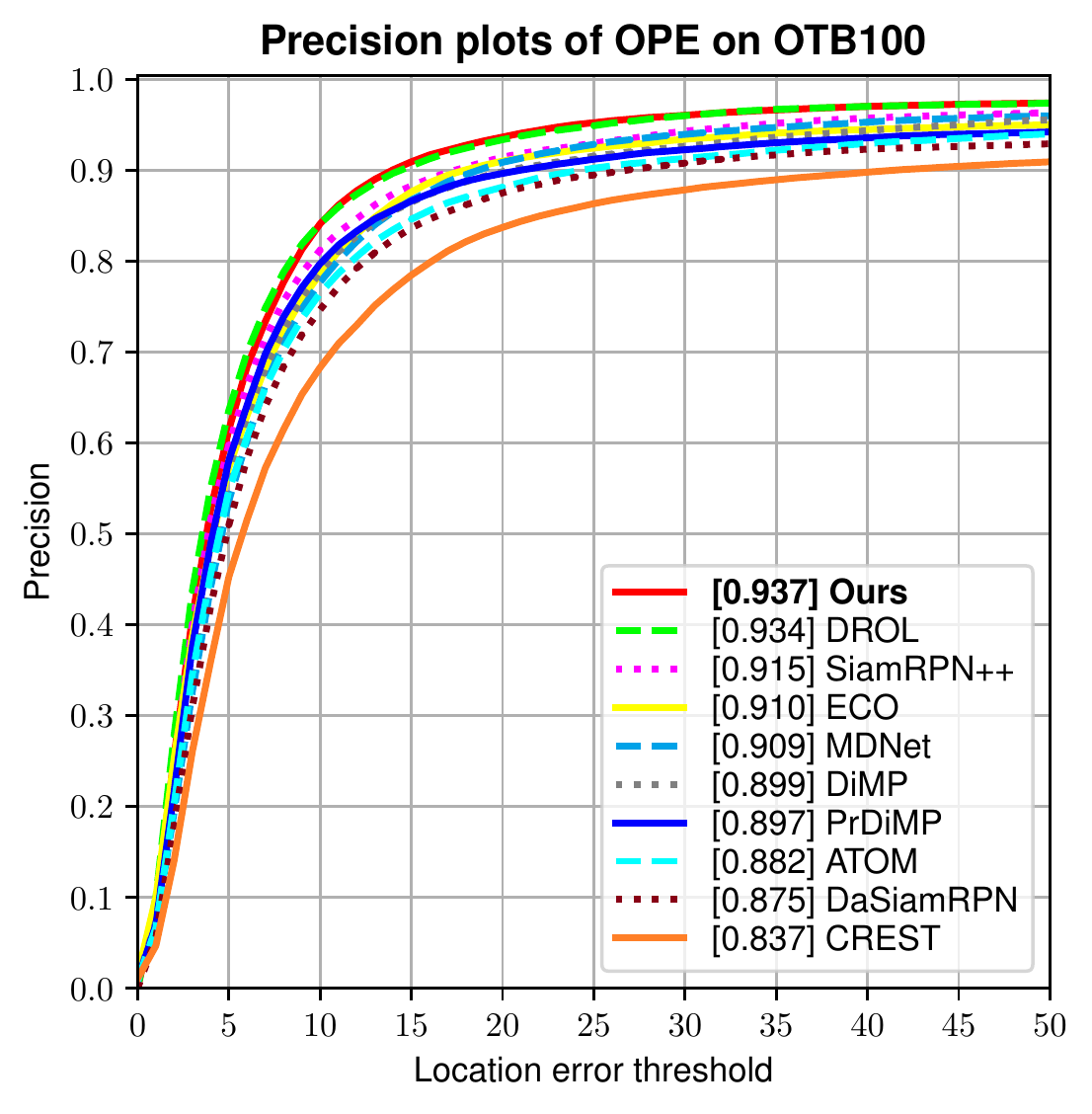}}
\subfigure{\includegraphics[width=0.33\linewidth,height=0.33\linewidth]{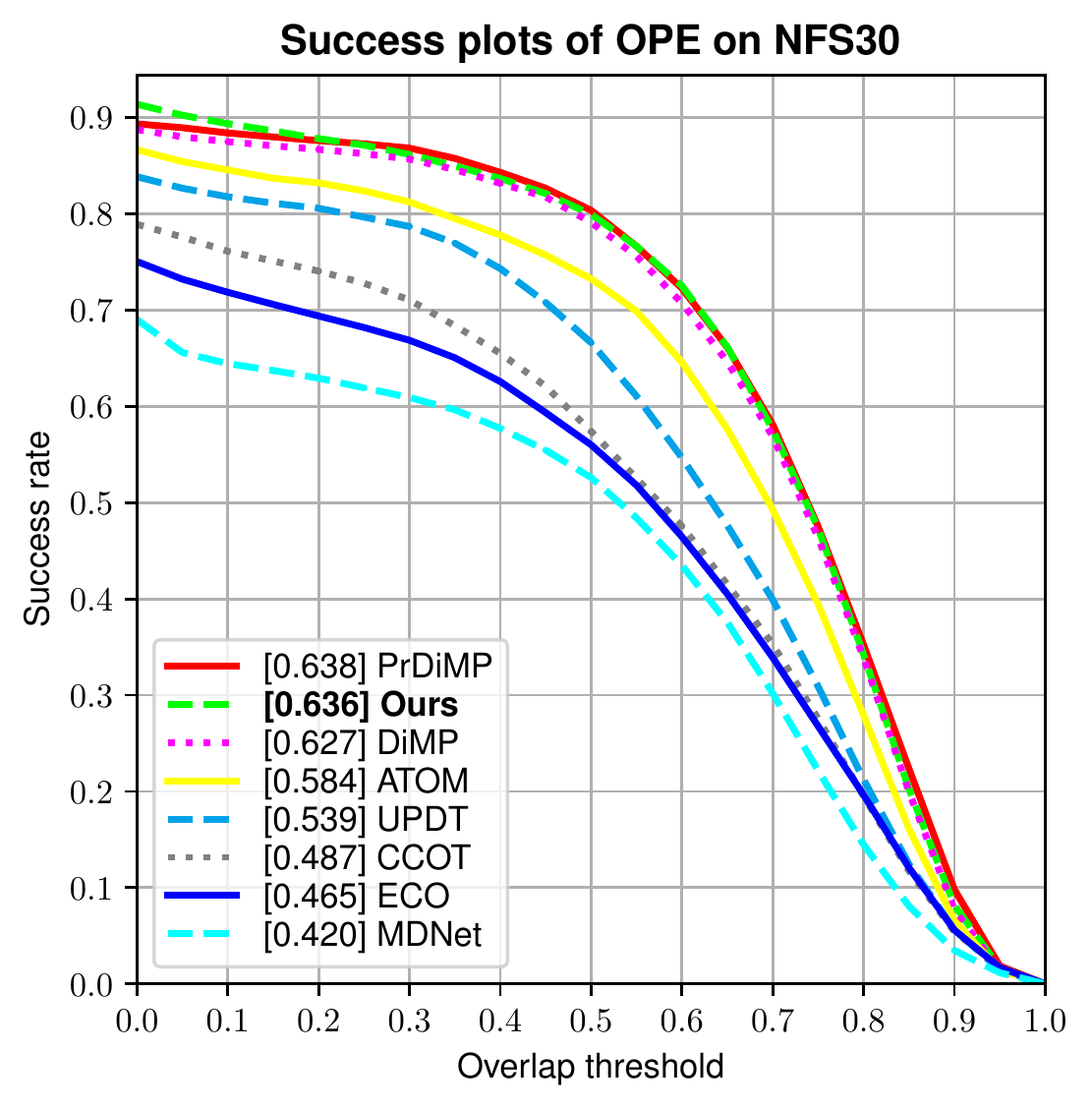}}\\
\subfigure{\includegraphics[width=0.33\linewidth,height=0.33\linewidth]{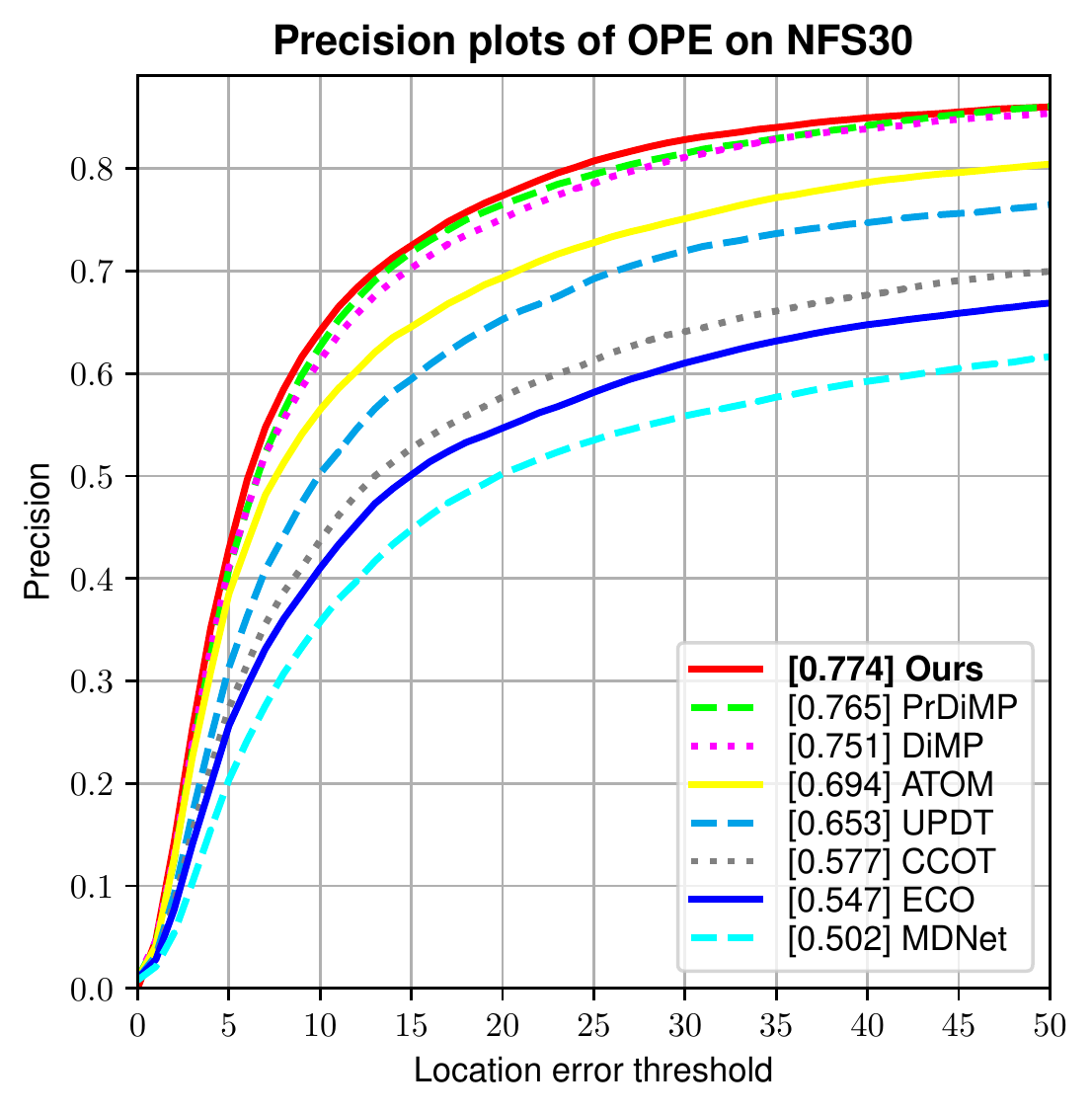}}
\subfigure{\includegraphics[width=0.33\linewidth,height=0.33\linewidth]{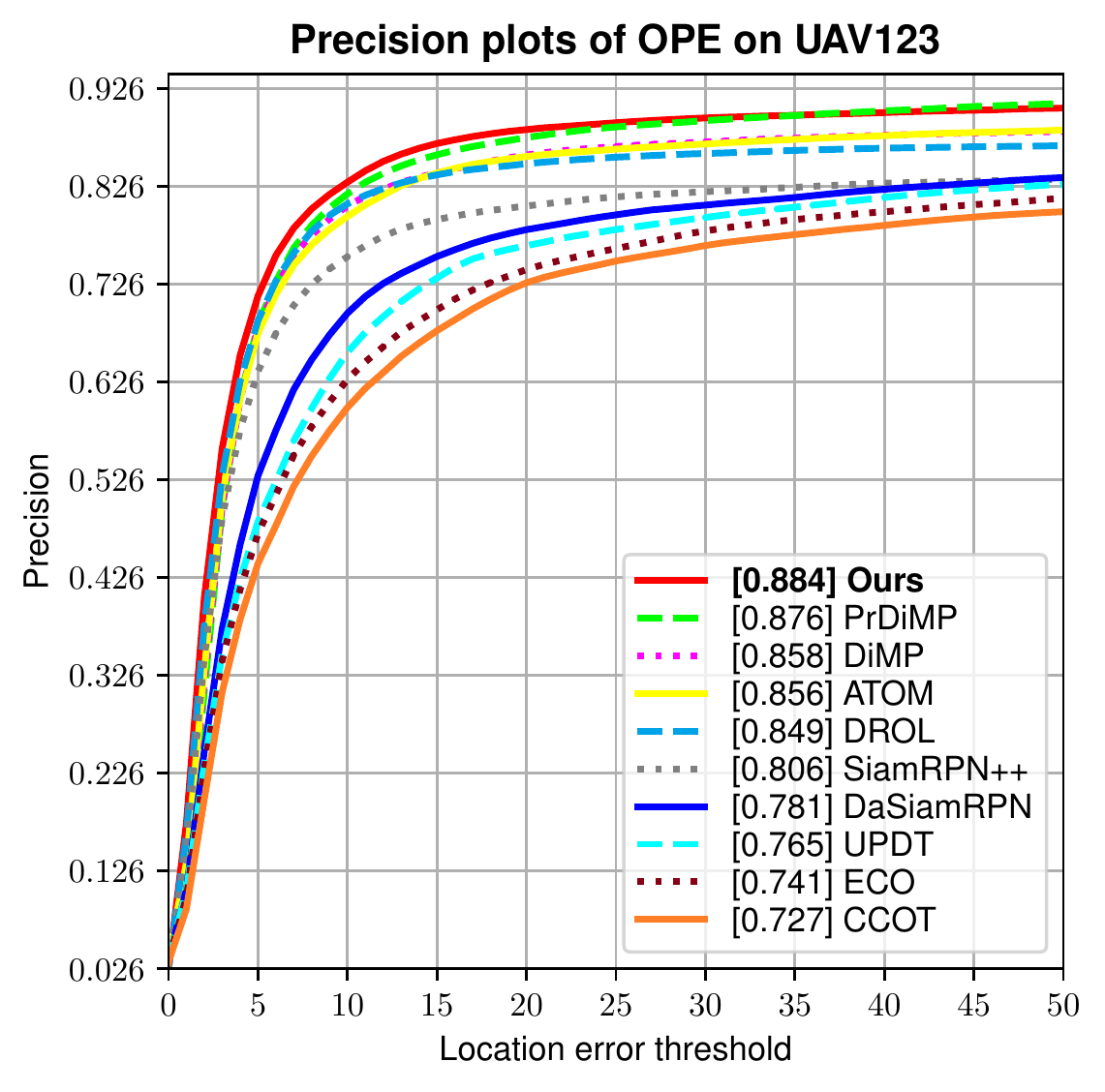}}
\subfigure{\includegraphics[width=0.33\linewidth,height=0.33\linewidth]{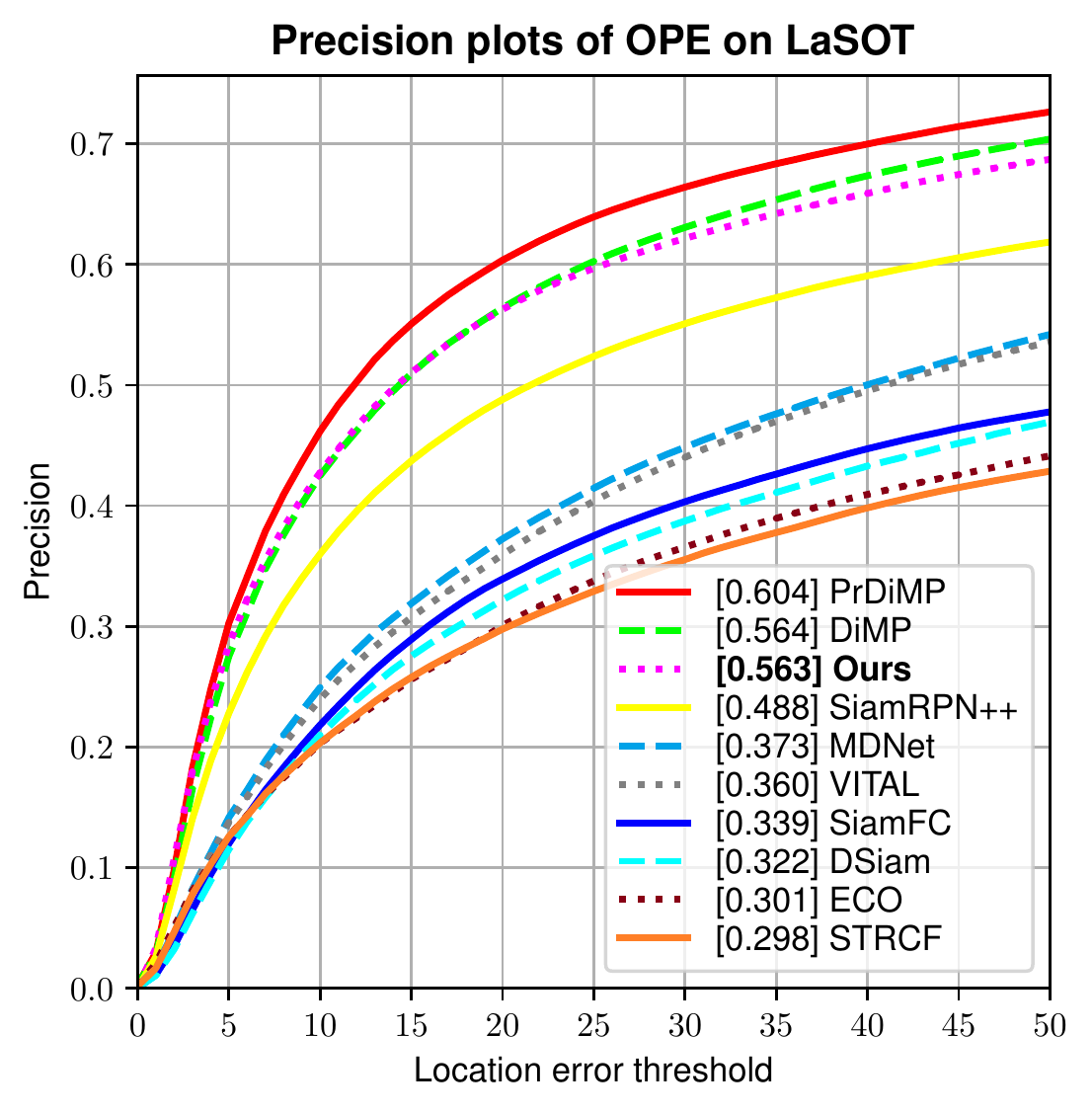}}
\vspace{-0.05cm}
\caption{Results in terms of AUC and Precision score on OTB2015, NFS, UAV123 and LaSOT.}
\vspace{-0.5cm}
\label{fig:detailed_results}
\end{figure*}

\section{Detailed Results}
\label{app:detailed_results}

We provided the detailed results in figures on OTB2015 \cite{otb2015}, NFS \cite{nfs}, UAV123 \cite{uav123}, and LaSOT \cite{lasot} datasets, as shown in Figure \ref{fig:detailed_results}. Our methods performs among the top methods for all the above datasets.

\begin{figure*}[!htbp]
\centering
\includegraphics[width=16cm]{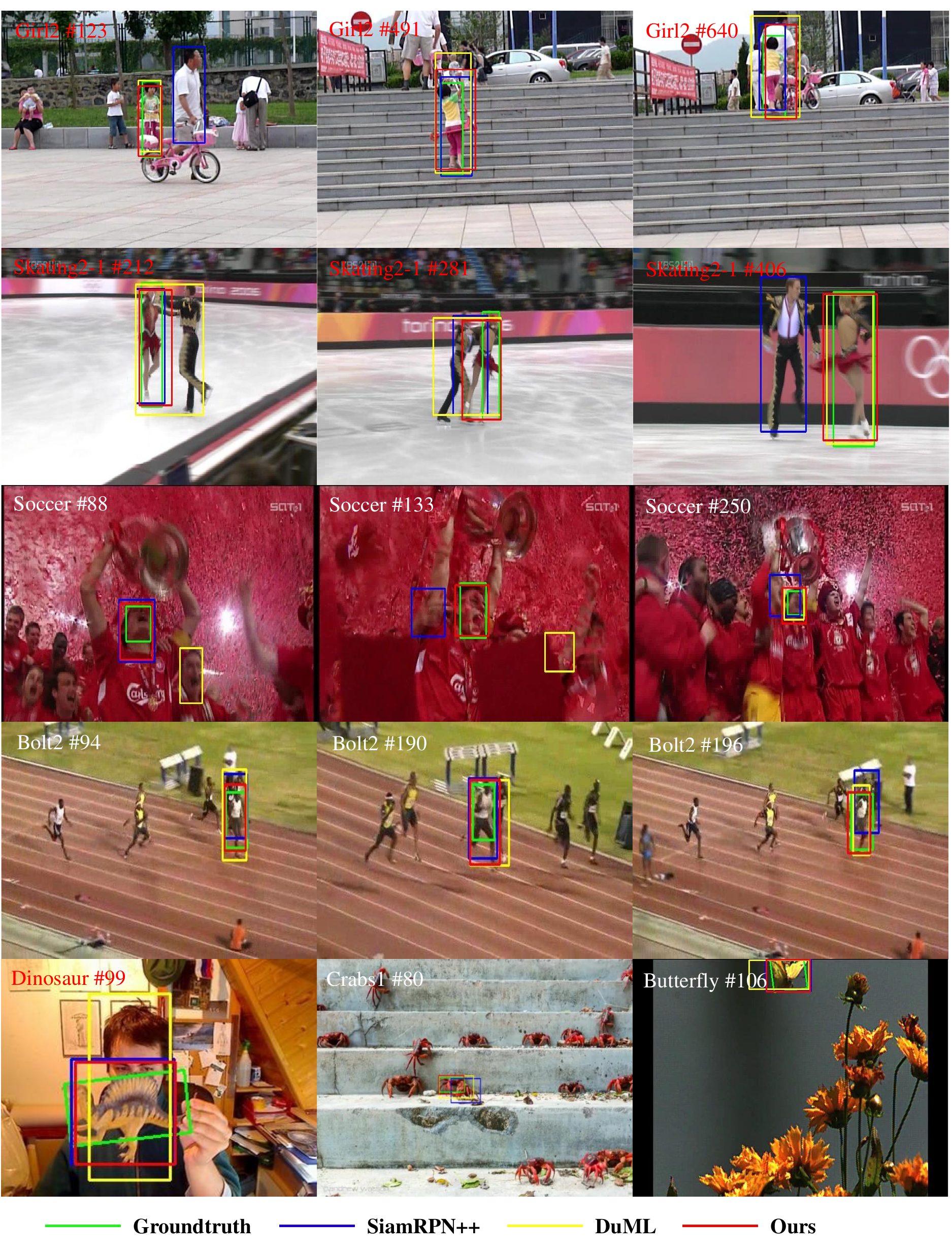}
\caption{Visualization of tracking results. Best viewed with color and zooming in. }
\label{fig:visual}
\vspace{-0.5cm}
\end{figure*}

\section{Visualization of the Voted Boxes} 
\label{app:iou_voting}

\begin{figure}[H]
\centering
\includegraphics[width=8.2cm]{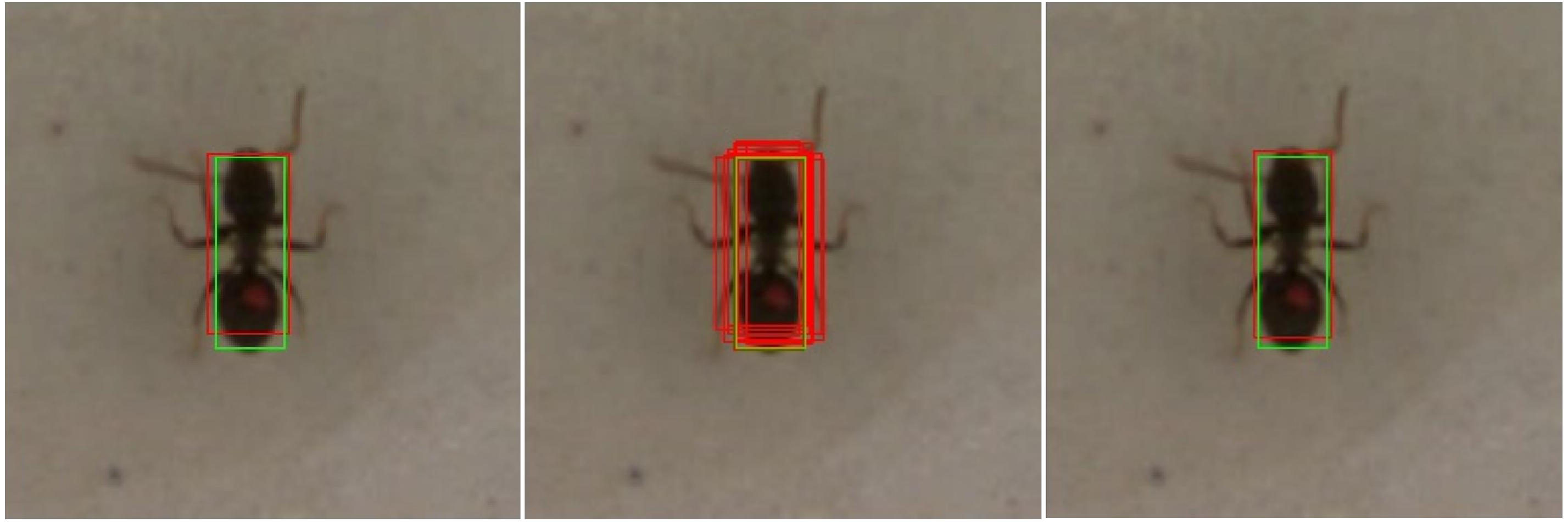}
\caption{Visualization of proposed score voting strategy. Best viewed with color.}
\label{fig:iouvoting}
\vspace{-0.5cm}
\end{figure}

To demonstrate the effectiveness of our proposed score voting strategy, we give a visualized results of voted boxes, shown in Figure \ref{fig:iouvoting}. \textcolor{green}{Green} denotes the groundtruth box and \textcolor{red}{red} denotes the predicted boxes. The bounding box from the leftmost figure is the predicted representation via direct regression. The middle figure showcases the chosen top-$10$ adjacent boxes according to their assigned voting weight. The final fused box representation is demonstrated in the rightmost figure. The box representation after score voting can approximate the groundtruth box more accurately, resulting in a more high-fidelity and precise representation of the target.

\section{Visualization of the Results} 
To validate the effectiveness of our proposed method, we showcase the visualized results on certain sequences from VOT2018 and OTB100. Our proposed method achieves a balance between robustness and accuracy. From the top $4$ columns of the Figure \ref{fig:visual}, compared with SiamRPN++ \cite{SiamRPN++}, we observe that our method fairly tackles the problem of target drift and inherits a strong discriminative ability from robust localization. Compared with DiMP \cite{DiMP}, our tracker can locate the target center better thus further yield a more accurate bounding box. From the last columns, Skating2-1 \#212 and Bolt2 \#94, we can see that DiMP tends to generate a larger bounding box when similar objects are present around the target. A contributing factor is that when the online module treats these adjacent pixels as positive samples, the target center may be shifted and the tracker tends to treat the cluster as the same object. Comparatively speaking, our tracker can locate the target center more precisely with accurate localization.

\end{document}